%% file: RSGDA_AISTATS.tex
\documentclass[twoside]{article}

\usepackage[preprint]{aistats2022_arxiv}

\usepackage[T1]{fontenc}
\usepackage{lmodern}
\usepackage{url}            
\usepackage{booktabs}       
\usepackage{amsfonts}       
\usepackage{nicefrac}       
\usepackage{microtype}      
\usepackage{graphicx}
\usepackage{natbib}
\usepackage{subcaption}
\usepackage{tikz-cd}
\usepackage{cuted}
\usepackage{lipsum}
\usepackage{multicol}

\makeatletter
\newcommand{\printfnsymbol}[1]{%
  \textsuperscript{\@fnsymbol{#1}}%
}

 \usepackage{authblk}    
\usepackage{geometry}   
\usepackage[utf8]{inputenc} 
\usepackage[pagebackref]{hyperref}       
\hypersetup{
  colorlinks=true,
  linkcolor=blue,
  filecolor=magenta,
  urlcolor=cyan,
  citecolor=blue
}
\usepackage{url}            
\usepackage{booktabs}       
\usepackage{amsfonts}       
\usepackage{nicefrac}       
\usepackage{microtype}      
\usepackage{wrapfig}
\usepackage{amsmath,amsthm,amssymb,mathtools,graphicx,enumitem,hyphenat,float}
\usepackage{bbm}
\usepackage{subcaption}
\usepackage{accents}
\usepackage{import}
\graphicspath{{../figures/}}
\setcitestyle{citesep={,}}

\usepackage{appendix}

\include{header}

\begin{document}

%

%

\twocolumn[

\aistatstitle{Randomized Stochastic Gradient Descent Ascent}

\aistatsauthor{ Othmane Sebbouh \And Marco Cuturi \And  Gabriel Peyré }

\aistatsaddress{ ENS, PSL, CNRS \\ CREST-ENSAE \And CREST-ENSAE \And ENS, PSL, CNRS } ]

\begin{abstract}
An increasing number of machine learning problems, such as robust or adversarial variants of existing algorithms, require minimizing a loss function that is itself defined as a maximum. Carrying a loop of stochastic gradient ascent (SGA) steps on the (inner) maximization problem, followed by an SGD step on the (outer) minimization, is known as Epoch Stochastic Gradient \textit{Descent Ascent} (ESGDA). While successful in practice, the theoretical analysis of ESGDA remains challenging, with no clear guidance on choices for the inner loop size nor on the interplay between inner/outer step sizes. We propose RSGDA (Randomized SGDA), a variant of ESGDA with stochastic loop size with a simpler theoretical analysis. RSGDA comes with the first (among SGDA algorithms) almost sure convergence rates when used on nonconvex min/strongly-concave max settings. RSGDA can be parameterized using optimal loop sizes that guarantee the best convergence rates known to hold for SGDA. We test RSGDA on toy and larger scale problems, using distributionally robust optimization and single-cell data matching using optimal transport as a testbed.
\end{abstract}

\input{sections/intro.tex}

\input{sections/background.tex}

\input{sections/method.tex}
\input{sections/applications.tex}
\input{sections/conclusion}
\input{sections/acknowledgements}


\bibliographystyle{apalike}
\bibliography{biblio}

\medskip


\clearpage

\appendix

\onecolumn 

\newpage

\input{sections/appendix.tex}

\end{document}

%% file: header.tex
\usepackage{wrapfig}

\usepackage{algorithm,algpseudocode}
\algdef{SE}[SUBALG]{Indent}{EndIndent}{}{\algorithmicend\ }%
\algtext*{Indent}
\algtext*{EndIndent}
\algblockdefx{Sync}{EndSync}[1]{\textbf{sync} #1}{\textbf{end sync}}
\usepackage{appendix}

\usepackage{amsmath}
\usepackage{cleveref}
\crefformat{section}{\S#2#1#3} 
\crefformat{subsection}{\S#2#1#3}
\crefformat{subsubsection}{\S#2#1#3}
\usepackage{autonum}

\usepackage{xcolor}
\usepackage{bm}

\usepackage[labelfont=bf, skip=1pt]{caption}
\graphicspath{{../figures}}
\usepackage{graphicx}
\newcommand\ec[2][]{\ensuremath{\mathbb{E}_{#1} \left[#2\right]}}
\newcommand\eczmu[2][]{\ensuremath{\mathbb{E}_{z \sim \mu} \left[#2\right]}}

\newcommand{\EE}[2]{\mathbb{E}_{#1}\left[#2\right] }
\newcommand\ecn[2][]{\ec[#1]{\norm{#2}^2}}

\newcommand\br[1]{\left ( #1 \right )}

\newcommand{\N}{\mathbb{N}}

\newcommand{\R}{\mathbb{R}}

\newcommand{\cD}{\mathcal{D}}
\newcommand{\cE}{\mathcal{E}}
\newcommand{\cF}{\mathcal{F}}
\newcommand{\cP}{\mathcal{P}}

\newcommand{\cL}{\mathcal{L}}
\newcommand{\cO}{\mathcal{O}}
\newcommand{\cV}{\mathcal{V}}
\newcommand{\cW}{\mathcal{W}}
\newcommand{\cX}{\mathcal{X}}
\newcommand{\cY}{\mathcal{Y}}
\newcommand{\cZ}{\mathcal{Z}}

\newcommand{\bnu}{\bm{\nu}}

\DeclareMathOperator*{\argmin}{\arg\!\min}
\DeclareMathOperator*{\argmax}{\arg\!\max}

\newcommand{\diag}{\mathrm{diag }}

\newcommand{\norm}[1]{\left\lVert#1\right\rVert}
\newcommand{\supnorm}[1]{{\left\lVert#1\right\rVert}_{\infty}}

\newcommand{\sqn}[1]{{\left\lVert#1\right\rVert}^2}

\newcommand{\abs}[1]{\left\lvert#1\right\rvert}

\newcommand{\eqdef}{\overset{\text{def}}{=}}
\newcommand{\ones}{{\bf 1}}
\newcommand*\diff{\mathop{}\!\mathrm{d}}
\makeatletter
\newsavebox\myboxA
\newsavebox\myboxB
\newlength\mylenA
\def\code#1{\texttt{#1}}

\usepackage{tcolorbox}
\usepackage{pifont}
\definecolor{mydarkgreen}{RGB}{39,130,67}

\definecolor{mydarkred}{RGB}{192,47,25}

\newcommand*\overbar[2][0.75]{%
    \sbox{\myboxA}{$\m@th#2$}%
    \setbox\myboxB\null
    \ht\myboxB=\ht\myboxA%
    \dp\myboxB=\dp\myboxA%
    \wd\myboxB=#1\wd\myboxA
    \sbox\myboxB{$\m@th\overline{\copy\myboxB}$}
    \setlength\mylenA{\the\wd\myboxA}
    \addtolength\mylenA{-\the\wd\myboxB}%
    \ifdim\wd\myboxB<\wd\myboxA%
       \rlap{\hskip 0.5\mylenA\usebox\myboxB}{\usebox\myboxA}%
    \else
        \hskip -0.5\mylenA\rlap{\usebox\myboxA}{\hskip 0.5\mylenA\usebox\myboxB}%
    \fi}
\makeatother

\usepackage[colorinlistoftodos,bordercolor=orange,backgroundcolor=orange!20,linecolor=orange,textsize=scriptsize]{todonotes}

\usepackage{thmtools}

\declaretheorem[within=section]{definition}

\declaretheorem[sibling=definition]{proposition}
\declaretheorem[sibling=definition]{assumption}
\declaretheorem[sibling=definition]{corollary}

\declaretheorem[sibling=definition]{lemma}

\declaretheorem[sibling=definition]{remark}

%% file: sections/intro.tex
\section{Introduction}
\textbf{Min-Max problems in ML.} Consider the following stochastic min-max optimization problem:
\begin{gather}\label{eq:main_problem}
    \min_{\theta \in \R^d} \phi(\theta)\,, \;\;\text{where}\;  \phi(\theta) \eqdef \max_{v \in \cV} F(\theta, v),\\
    \text{and} \quad F(\theta, v) = \EE{z}{f(\theta, v; z)},
\end{gather}
with $\cV \subseteq \R^n$. Problems such as \eqref{eq:main_problem} appear in the estimation of generative models \citep{goodfellow2014generative}, reinforcement learning \citep{dai2018sbeed}, online learning \citep{cesa2006prediction}, and in many other domains including mathematics and economics \citep{von2007theory, bacsar1998dynamic} (see \cite{nouiehed2019solving} and references therein).


\textbf{Gradient Descent-Ascent algorithms.} Most machine learning applications resort to stochastic gradient methods to solve~\eqref{eq:main_problem}. These methods, which we refer to as \textit{Gradient Descent Ascent (GDA) algorithms}, alternate between possibly many ascent steps in $v$ to increase $F$, with a descent step along a stochastic gradient direction to decrease $\phi$. Within that space alone, several algorithms have been proposed, some of them taking advantage of assumptions on the properties of $F$. The setting that has received the most attention is by far that where $F$ is convex-concave \citep{sion1958general, korpelevich1976extragradient, nemirovski2004prox, nedic2009subgradient, azizian2020tight}. Only recently has the application of GDA to non-convex functions been thoroughly analyzed from a theoretical perspective, resulting in a flurry of papers \citep{namkoong2016stochastic, sinha2018certifying, rafique2018non, grnarova2017online, lu2020hybrid, nouiehed2019, thekumparampil2019efficient, kong2019accelerated, jin2020local, zhang2020single}. These recent works are particularly relevant to modern machine learning problems, where the \textit{min} problem in \eqref{eq:main_problem} often require optimizing the parameters of models using a non-convex loss $\phi$. 

\textbf{Non-convex strongly concave Optimization.} 
We consider in this work the setting where $F$ is nonconvex in $\theta$, yet strongly concave and smooth in $v$. This setting is practically relevant: since it appears in Temporal Difference learning \citep{dai2018sbeed}, robust optimization \citep{sinha2018certifying}, or entropic optimal transport \citep{cuturi2013sinkhorn}. It is also theoretically appealing, since it has been shown that GDA algorithms achieve a complexity similar to single-variable minimization, up to a factor depending on the conditioning of the problem \citep{sanjabi2018solving, nouiehed2019, lin2020, qiu2020single, huang2020accelerated}. For example, compared to the nonconvex-concave and nonconvex-nonconcave settings, one does not need to assume that the set $\cV$ in \eqref{eq:main_problem} is bounded in order to ensure convergence.

In practice, one of the most widely used algorithms to solve \eqref{eq:main_problem} is epoch stochastic gradient descent ascent \citep{goodfellow2014generative, sanjabi2018convergence, lin2020, jin2020local, nouiehed2019, sanjabi2018solving, houdard2021existence}, where we make an arbitrary fixed number of stochastic gradient ascent steps followed by a gradient descent step. Unfortunately, there is very little understanding of the theoretical justifications behind this method, leaving practitioners in the dark as to what guarantees they might expect from their parameter settings.


\textbf{Our Contributions.}  In this work, we aim to quantify -- in the form of convergence rates and suggested parameter settings -- how much the practical choices are justified by theory. To this end, we propose Randomized Stochastic Gradient Descent Ascent (RSGDA), a randomization of Epoch Stochastic Gradient descent ascent (ESGDA) which is more amenable to theoretical analysis and performs very similarly to ESGDA in practice. From a theoretical perspective, we show that RSGDA enjoys the best known convergence rates, which are also verified by one-step stochastic gradient descent ascent \citep{lin2020} and other variants \citep{huang2020accelerated, qiu2020single}, and we demonstrate how the step sizes and number of gradient ascent steps should be set in order to retain these convergence guarantees. We evaluate our suggested parameter settings on problems from robust optimization and optimal transport: we consider $(i)$ the problem of distributionally robust optimization \citep{sinha2018certifying}, where we aim to learn a classifier which is robust to adversarial inputs, and $(ii)$ the problem of single-cell data matching using regularized optimal transport \citep{schiebinger2019optimal, stark2020scim, cuturi2013sinkhorn}.

%% file: sections/background.tex
\section{Background}
We present in this section relevant assumptions for analysis, and briefly review state of the art results.

\subsection{Assumptions}
\paragraph{Smoothness and strong concavity.} Throughout the paper, we assume that $F$ is smooth in both variables, and strongly concave in the second variable.
\begin{assumption}\label{asm:smooth_str_concave}
We assume that F is $L$-smooth on $\R^d$ and $v \mapsto F(\theta, v)$ is $\mu$-strongly concave on $\cV$.
\end{assumption}
The smoothness and strong concavity of $F$ ensure that the function $\phi$ defined in \eqref{eq:main_problem} is smooth as well.
\begin{lemma}[Lemma 4.3 in \cite{lin2020}]\label{lem:general_smooth_lip}
Let Assumption \ref{asm:smooth_str_concave} hold. Let $\kappa \eqdef \frac{L}{\mu}$. Define ${v^* : \R^d \mapsto \R^n}$ by $v^*(\theta) = \argmax_{v \in \R^n} F(\theta, v)$ for all $\theta \in \R^d$. Then, $v^*$ is $\kappa$-lipschitz and $\phi$ is $2\kappa L$-smooth.
\end{lemma}

\paragraph{Assumptions on noise.} We assume that the stochastic gradient in $\theta$ has bounded variance, which is standard in non-convex stochastic optimization.
\begin{assumption}\label{asm:noise_theta}
There exists $\sigma^2 > 0$ such that for all $(\theta, v) \in \R^d \times \cV$,
\begin{align}
    \EE{z}{\sqn{\nabla_\theta f(\theta, v; z) -  \nabla_\theta F(\theta, v)}} \leq \sigma^2.
\end{align}
\end{assumption}

For the stochastic gradient in $v$, we only assume that the noise is \textit{finite at the maximizer}.
\begin{assumption}\label{asm:noise_v}
Define $v^*(\theta) \eqdef \underset{v \in \R^n}{\argmax} f(\theta, v)$. Let $\tilde{\sigma}^2 \eqdef \EE{z}{\sqn{\nabla_v f(\theta, v^{*}(\theta); z)}} < \infty$.
\end{assumption}

\subsection{Review of Gradient Descent Ascent Methods}

\textbf{Definitions.} Let $\epsilon, \delta > 0$. We call $\theta \in \R^d$ an $\epsilon$-\textit{approximate stationary point} if $\norm{\nabla \phi(\theta)} \leq \epsilon.$ We call $v$ a $\delta$-\textit{approximate maximizer} if for some $\theta \in \R^d$ we have  $\phi(\theta) - F(\theta, v) \leq \delta$. 

\textbf{SGDmax, SGDA and ESGDA.} One of the most analyzed algorithms for solving min-max problems such as \eqref{eq:main_problem} is (S)GDmax \citep{sanjabi2018convergence, lin2020, jin2020local, nouiehed2019, sanjabi2018solving, houdard2021existence}, where at each iteration we make the number of (stochastic) gradient ascent steps necessary to reach a $\delta$-approximate maximizer $v_{k+1}$, before making a descent step using the gradient $\nabla_\theta f(\theta_k, v_k)$. See Alg.~\ref{alg:all-SGDA}, \textbf{SGDmax}. To reach an $\epsilon$-stationary point $\theta_k$, this algorithm requires $\cO\br{\log\br{1/\delta}\kappa^2\epsilon^{-2}}$ (resp. $\cO\br{\log\br{1/\delta}\kappa^3\epsilon^{-4}}$) total gradient computations in the deterministic (resp. stochastic) setting \citep{lin2020}. In practice, however, because SGDmax involves a subroutine where we need to ensure that we reach an approximate maximizer, this algorithm is rarely implemented. It is instead approximated by ESGDA.

\begin{algorithm}
\caption{SGDmax/SGDA/ESGDA}
\label{alg:all-SGDA}
\begin{algorithmic}
    \State \textbf{Inputs:} step sizes $\alpha$ and $\eta$, loop size $m$, max-oracle accuracy $\delta$
    \For{$k=0, 1, 2, \dots,$}
    \State \textbf{\underline{SGDmax}: }
    \Indent 
    \State Find $v_{k+1}$ s.t. ${\EE{k}{F(\theta_k, v_{k+1})} \geq \phi(\theta_k) + \delta}$
    \EndIndent
    \State \textbf{\underline{SGDA}: }
    \Indent 
    \State Sample $z_k^{\prime} \sim \cD$
    \State $v_{k+1} = \Pi_{\cV}\left(v_k + \eta \nabla_v f(\theta_k, v_k; z_k^{\prime})\right)$
    \EndIndent
    \State \textbf{\underline{ESGDA}: }
    \Indent
    \For{$t=0,1, 2, \dots, m - 1$}
    \State Sample $z_k^t \sim \cD$
    \State $v_k^{t+1} = \Pi_{\cV}\left( v_k^t + \eta \nabla_v f(\theta_k, v_k^t; z_k^t) \right)$
    \EndFor
    \State $v_{k+1} = v^0_{k+1} = v_k^m$
    \EndIndent
    \State Sample $z_k \sim \cD$
    \State $\theta_{k+1} = \theta_k - \alpha \nabla_\theta f(\theta_k, v_{k+1}; z_k)$
    \EndFor
\end{algorithmic}
\end{algorithm}

At the other end of the spectrum is one-step Gradient Descent Ascent \citep{lin2020, chen2020proximal} (also referred to as GDA in the literature), in which one ascent step is followed by one descent step. See Alg. \ref{alg:all-SGDA}, \textbf{SGDA}. This algorithm has two important advantages compared with (S)GDmax: (a) it is simple: it doesn't require any inner ascent loop or stopping criteria, (b) it has better convergence rates: $\cO\br{\kappa^2\epsilon^{-2}}$ (resp. $\cO\br{\kappa^3\epsilon^{-4}}$)  in the deterministic (resp. stochastic) setting.

In practice, instead of choosing a precision $\delta$ or running SGDA, one popular choice \citep{goodfellow2014generative, sinha2018certifying, houdard2021existence} is using epoch stochastic gradient descent ascent (ESGDA), where we make a fixed number of ascent steps on $v$ followed by a descent step on $\theta$. See Alg.~\ref{alg:all-SGDA}, \textbf{ESGDA}. The goal of the ascent steps is to have a good enough approximation of $v^*(\theta_k)$, and hence of the gradient $\nabla \phi(\theta_k)$, in order to make a descent step on $\phi$. Despite its popularity, we know little about the theoretical properties of ESGDA in the nonconvex-strongly concave setting. \cite{yan2020optimal} studies a version of ESGDA with an \textit{iteration dependent} number of gradient ascent steps. The problem they consider -- $F$  weakly convex and strongly concave -- is inherently harder than the smooth nonconvex-strongly concave problem. The complexity of their method for finding a nearly stationary point is $\cO\br{\epsilon^{-4}}$, but to reach a stationary point, they still need $\cO\br{\epsilon^{-6}}$ iterations, even when their results are specialized to the smooth setting (See Prop. 4.11 in \cite{lin2020} for the relation between stationarity and near stationarity). \cite{chen2021tighter} consider the harder problem of stochastic nested optimization (which includes bilevel and min-max optimization). They devise an algorithm (ALSET) with $\cO\br{\epsilon^{-4}}$ complexity when the epoch size is $\Theta(\kappa)$. But they assume that their function and its hessian are both lipschitz, whereas the analysis of SGDA only requires that the gradients are lipschitz.

Despite the popularity of ESGDA, to the best of our knowledge, there is no theoretical analysis ensuring that this algorithm converges under the same conditions as SGDA and SGDmax.  In this work, we aim to support practical implementation choices of ESGDA with solid theory, and in turn suggest new parameter settings to further improve how ESGDA is implemented.

\textbf{Motivating RSGDA.} RSGDA is a randomized version of ESGDA with a \textit{stochastic} loop size. Empirically, RSGDA performs similarly to ESGDA (Fig.~\ref{fig:mnist_proba_loop}), but its theoretical analysis is simpler, thanks to its inner-loop free structure. This is showcased in Prop.~\ref{prop:conv_rsgda}, our central result, a descent inequality (up to additional noise terms), from which we draw several conclusions. First, we derive \textit{almost sure} convergence rates for RSGDA (Cor.~\ref{cor:as_conv_rsgda}). Then, we show that, like SGDA, RSGDA enjoys the best known convergence rates in expectation among stochastic gradient descent ascent algorithms (Cor.~\ref{cor:rsgda_big_batch}). Importantly, for each of our convergence results, we determine \textit{(i)} a range of descent step probabilities $p$ that guarantees the best possible convergence rate, and \textit{(ii)} how the step sizes should be set depending on $p$.

%% file: sections/method.tex
\section{Randomized Stochastic Gradient Descent Ascent}\label{sec:rsgda}
We now introduce RSGDA (Alg.~\ref{alg:RSGDA}) in detail. At each iteration $k$, we toss a coin: if it lands heads (with probability $1 - p$), we keep $\theta_k$ fixed and make an ascent step on $v_k$ along the stochastic gradient $\nabla_v f(\theta_k, v_k; z_k)$, where $z_k \sim \cD$ (or a projected ascent step if $\cV \neq \R^n$); if it lands tails (with probability $p$), we keep $v_k$ fixed and make a descent step along the stochastic gradient $\nabla_\theta f(\theta_k, v_k; z_k)$. The algorithm can also be seen as a version of ESGDA where the size of the inner loop $m$ is stochastic and equal to $1/p - 1$ in expectation.

\begin{algorithm}[h]
\caption{Randomized SGDA}
\label{alg:RSGDA}
\begin{algorithmic}
    \State \textbf{Inputs:} step sizes $(\alpha_k)_k$ and $(\eta_k)_k$, $p \in (0,1)$.
    \State \textbf{Initialisation:} $v_0 \in \R^n, \theta_0 \in \R^d$
    \For{$k=0, 1, 2, \dots,$}
    \State Sample $z_k \sim \cD$
    \State $\theta_k^+ = \theta_k - \alpha_k \nabla_\theta f(\theta_k, v_k; z_k)$
    \State $v_k^+ = \Pi_{\cV}\br{v_k + \eta_k \nabla_v f(\theta_k, v_k; z_k)})$
    \begin{eqnarray*} (\theta_{k+1}, v_{k+1})
    =\left\{
    \begin{array}{ll}
        (\theta_k^+, \; v_k) & \mbox{\textbf{w. p.} }p \\
        (\theta_k \;, v_k^+) & \mbox{\textbf{w. p.} } 1-p
    \end{array} \right.
    \end{eqnarray*}
    \EndFor
\end{algorithmic}
\end{algorithm}

From a stochastic optimization perspective, this randomization trick is reminiscent of the way Loopless SVRG (Stochastic Variance Reduced Gradient) \citep{hofmann2015variance, kovalev2020don} avoids using the inner loop of the original SVRG method \citep{johnson2013accelerating}. Like for Loopless SVRG and SVRG, RSGDA results in a much simpler analysis than for ESGDA.

This is apparent through Prop.~\ref{prop:conv_rsgda}, in which we establish a one-step recurrence inequality which is central to deriving our convergence results.

\begin{proposition}\label{prop:conv_rsgda}
Consider the iterates of Alg.~\ref{alg:RSGDA}. Let Assumption~\ref{asm:smooth_str_concave} hold. Define for all $k \in \N$,
\begin{align}
\cD_k &= \phi(\theta_{k}) - \min_{\theta \in \R^d}\phi(\theta), \quad r_k = \sqn{v^*(\theta_k) - v_k}, \\
& \mbox{and} \quad \cE_k = \cD_k + \kappa L \frac{p\alpha_{k}}{(1-p)\eta_k}r_k.
\end{align}
Let $\br{\alpha_k}_k$ and $\br{\eta_k}_k$ be two positive decreasing sequences such that $\br{\frac{\alpha_k}{\eta_k}}_k$ is decreasing as well, with $\eta_k \leq \frac{1}{2L}$ and $\alpha_k \leq \frac{(1-p)\eta_k}{4\kappa^2\sqrt{p\br{2p + (1-p)\eta_k\mu}}}$. Then,
\begin{align}
    &p\alpha_k \sqn{\nabla \phi(\theta_k)} + 2\EE{k}{\cE_{k+1}} \leq 2\cE_{k} + 4\eta_k p \alpha_k \kappa L\tilde{\sigma}^2\\
    &+ 2\sigma^2\br{ p\alpha_k^2\kappa L + \frac{p^2\br{2p + \br{1-p}\eta_k\mu}\alpha_k^3\kappa^4}{(1-p)^2\eta_k^2}}. \label{eq:conv_reccurrence_rsgda}
\end{align}
\end{proposition}

\begin{figure}
    \centering
    \includegraphics[scale=0.65]{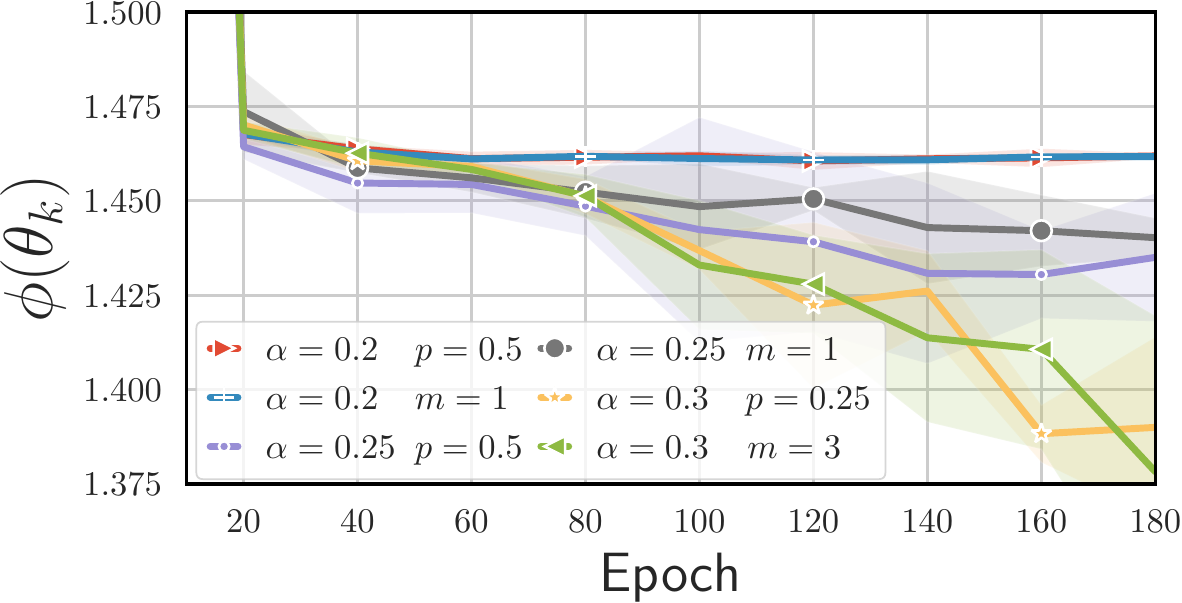}
    \caption{We use RSGDA (Alg. \ref{alg:RSGDA}) and ESGDA (Alg. \ref{alg:all-SGDA}) on the minmax problem described in \cref{sec:dro}. This figure shows that ESGDA with a given loop size $m$ and RSGDA with $p= 1/(m+1)$ perform similarly. The objective $\phi$ is defined in~\eqref{eq:drw}.}
    \label{fig:mnist_proba_loop}
    \vspace{-1.5em}
\end{figure}

\textbf{Almost sure convergence rates.} Establishing the recurrence inequality of Prop.~\ref{prop:conv_rsgda} is key to deriving almost sure convergence results in stochastic optimization \citep{bertsekas2000gradient, gadat2018stochastic, sebbouh2021almost}, where one can directly apply the Robbins-Siegmund theorem \citep{Robbins71}. In contrast, it is unclear how almost sure convergence rates can be derived for {ESGDA}.

Similar to what was done in \cite{sebbouh2021almost} for SGD, we can use this lemma and Prop.~\ref{prop:conv_rsgda} to derive 'small-o' almost sure convergence \textit{rates} for RSGDA. To the best of our knowledge, these are the first such rates for a gradient descent ascent algorithm.

\begin{corollary}\label{cor:as_conv_rsgda}
Consider the setting of Prop.~\ref{prop:conv_rsgda}. We have that 
\begin{align}
\min_{t=0,\dots,k-1} \sqn{\nabla \phi(\theta_t)} = o\br{\frac{1}{\sum_{t=0}^{k-1}\alpha_t}}  \text{a.s}. \label{eq:almost_sure_rate_general}
\end{align}
as long as the step sizes $\alpha_k$ and $\eta_k$ verify
\begin{gather}
\begin{aligned}\label{eq:conditions_stepsizes_as_rsgda}
    \sum_k \alpha_k = \infty, \; \sum_k \alpha_k^2\sigma^2 < \infty,\\
     \sum_k \eta_k \alpha_k\tilde{\sigma}^2 < \infty, \; \sum_k \frac{\alpha_k^3}{\eta_k^2}\sigma^2 < \infty. 
\end{aligned}
\end{gather}

Let $\zeta > 0$ and take for all $k \in \N$,
    $\eta_k = \frac{1}{2L(k+1)^{2/5+\zeta}}$ and $\alpha_k = \frac{1-p}{
    2\sqrt{p\br{2p + (1-p)\eta_k\mu}}}\frac{\eta_k}{\kappa^2(k+1)^{1/5}}$,
we have that the step sizes verify \eqref{eq:conditions_stepsizes_as_rsgda} and
\begin{align}
    \min_{t=0,\dots,k-1} \sqn{\nabla \phi(\theta_t)} = o\br{k^{-\frac{2}{5} + \zeta}} \; \mbox{almost surely}.
\end{align}
\end{corollary}
\begin{remark}
Using the same analysis, it is also possible to show that using the exact gradients, Randomized GDA converges \textit{almost surely} at a $o(k^{-1})$ rate. Indeed, in that setting, $\sigma^2 = \tilde{\sigma}^2 = 0$ and any constant step sizes $\alpha, \eta > 0$ verify conditions \eqref{eq:conditions_stepsizes_as_rsgda}. Substituting in \eqref{eq:almost_sure_rate_general} gives the desired rate. 
\end{remark}

\textbf{Convergence rates in expectation.} In the remainder of this section, we establish convergence rates in expectation for Randomized GDA (where we use the exact gradient at each iteration) and RSGDA (where we use stochastic gradients). The rates we recover are similar to the best that can be derived for GDA and SGDA \citep{lin2020}. The interest of the results we present is that  $(i)$ we show that it is possible to do more gradient ascent steps and retain the same complexity as SGDA and we quantify this by determining a range of values of $p$ for which we have this complexity, $(ii)$ we highlight the effect of changing the probability $p$ on the step sizes, and $(iii)$ we show that taking $p \geq 1/2$ doesn't result in a diverging algorithm; instead the convergence rate is worse by a factor which is an increasing function of $p$, so that the strategy of making moderately more descent steps than ascent steps is worth exploring depending on the application.

\subsection{Randomized GDA (RGDA)}
We first consider RGDA, a version of Alg.~\ref{alg:RSGDA} where we use the exact gradients, \textit{i.e.} where the only randomness in the algorithm comes from the coin tosses governed by the probability $p$. This is the randomized equivalent of Epoch GDA.
\begin{corollary}\label{cor:conv_rgda}
Consider the setting of Prop.~\ref{prop:conv_rsgda}. Let $\eta_k = \eta = \frac{1}{2L}$ and $\alpha = \frac{\br{1-p}}{4\kappa^2L}\frac{1}{\sqrt{p\br{2p + \frac{1-p}{2\kappa}}}}$. Then, for all $p \in \left[\frac{1}{\kappa}, \frac{1}{2}\right]$, we have that $$\min_{t=0,\dots,k-1} \ec{\sqn{\nabla \phi(\theta_t)}} = \cO\br{\kappa^2k^{-1} }.$$
Hence, finding an $\epsilon$-stationary point requires at most $\cO(\kappa^2\epsilon^{-2})$ iterations for all $p \in \left[\frac{1}{\kappa}. \frac{1}{2}\right]$.
\end{corollary}
\textbf{Remark.} The value $1/2$ of the upper bound on $p$ is arbitrary, and any constant higher than $1/2$ but independent of $\kappa$ still ensures the same complexity. 

We draw two important insights from this corollary:

\textbf{(a) Flexibility to the choice of $p$.} RGDA admits a range of values of $p$ for which we have the same order of complexity. In particular, with $p = 1/2$, we recover the same convergence rate for GDA which was derived in \cite{lin2020}, with better constants, as highlighted in the appendix.

\textbf{(b) Suggested choice of step sizes.} It is known from \cite{lin2020} that the step sizes $\alpha$ and $\eta$ need to verify the following relation: $\alpha = \Theta\br{\eta/\kappa^2}$. How should the step sizes change when we make more ascent steps? Since $p \in [1/\kappa, 1/2]$ gives the tightest convergence rate, let us consider the case where $p$ is in this range. In this setting, Cor.~\ref{cor:conv_rgda} shows that the relation should be $\alpha = \Theta\br{\eta/\mathbf{p}\kappa^2}$. Hence, there is a linear relation between the descent step size and $1/p$: the lower $p$, \textit{i.e.} the more ascent steps we make, the higher $\alpha$ should be.

We can draw similar conclusions in the stochastic setting as well, but with different ranges for $p$ and different choices for the step sizes.

\subsection{Randomized SGDA (RSGDA)}
We now consider RSGDA, where we use stochastic gradients instead of deterministic ones. This is the randomized version of Epoch SGDA.
\paragraph{Decreasing step sizes.} As is the case for SGD, without additional structure on the objective $F$, we can only guarantee the \textit{anytime convergence} (without knowledge of the last iteration or the required precision) of RSGDA if we use decreasing step sizes.

\begin{corollary}\label{cor:exp_conv_rsgda}
Consider the setting of Prop.~\ref{prop:conv_rsgda}. Let $p = p \in (0,1)$, $\eta_k = \frac{1}{2L(k+1)^{2/5}}$ and $\alpha_k = \frac{1-p}{
    2\sqrt{p\br{2p + (1-p)\eta_k\mu}}}\frac{\eta_k}{\kappa^2(k+1)^{1/5}}$. Then,
 $$\min_{t=0,\dots,k-1} \sqn{\nabla \phi(\theta_t)} = \cO\br{\log(k)k^{-2/5}}.$$
\end{corollary}
The previous corollary presented the parameter settings that resulted in the best rate when using decreasing step sizes. We have used the $\cO$ notation for brevity even though the result is non-asymptotic. A complete version of this corollary, including the explicit bound, can be found in the appendix.

\paragraph{RSGDA with fixed step sizes.} In practice, it is often the case that stochastic gradient algorithms are implemented using a fixed step size.
\begin{corollary}\label{cor:rsgda_fixed_stepsizes}
Consider the setting of Prop.~\ref{prop:conv_rsgda}. Let $\epsilon > 0$. With suitable choices of step sizes, if $k \geq \Theta\br{\kappa^3\epsilon^{-5}}$, then $\min_{t=0,\dots,k-1} \ec{\norm{\nabla \phi(\theta_t)}} \leq \epsilon$.
Moreover, for all $p \in \left[\frac{\epsilon}{\kappa^2}, \frac{1}{2}\right]$, the complexity is still $\cO\br{\kappa^3\epsilon^{-5}}$.
\end{corollary}
We give the explicit values for the step sizes and the complexity in the appendix. In this corollary, we recover the complexity given without proof in~\cite{lin2020} when using $p = 1/2$, but we extend their result to a larger number of ascent steps (instead of a single one). Note that the step sizes rely on some quantities that are generally impossible to obtain, like $\sigma$. As is the case for SGD, without knowledge of this constant and using constant step sizes, it is possible to guarantee only sublinear convergence towards \textit{the neighborhood} of a solution (see e.g.~\citep{gower2019sgd}). 

\paragraph{RSGDA with large batch sizes.} The best known complexity for SGDA is $\cO\br{\kappa^3 \epsilon^{-4}}$ \citep{lin2020}, and it is obtained using fixed step sizes and large minibatch sizes. In the following corollary, we show that the same rate can be recovered for RSGDA using large minibatch sizes. As done in the previous corollaries, we also give the range of values of $p$ which give the same rate. Following \cite{lin2020}, we assume that all stochastic gradients have bounded variance in this setting.

\begin{assumption}\label{asm:noise_large_batch}
There exists $\bar{\sigma}^2 > 0$ such that for all $(\theta, v) \in \R^d \times \cV$,
\begin{align}
    \EE{z}{\sqn{\nabla_\theta f(\theta, v; z) -  \nabla_\theta F(\theta, v)}} &\leq \bar{\sigma}^2\\
    \EE{z}{\sqn{\nabla_v f(\theta, v; z) -  \nabla_v F(\theta, v)}} &\leq \bar{\sigma}^2.
\end{align}
\end{assumption}

\begin{corollary}[Large minibatch sizes]\label{cor:rsgda_big_batch}
Consider the setting of Prop.~\ref{prop:conv_rsgda}. Let Assumption~\ref{asm:noise_large_batch} hold and choose $p \in [\frac{1}{\kappa}, \frac{1}{2}]$. Using the step sizes of Cor.~\ref{cor:conv_rgda} and a sufficiently large minibatch size, the total number of stochastic gradient evaluations to reach an $\epsilon$-stationary point is $\cO\br{\kappa^3\epsilon^{-4}}$.
\end{corollary}

\paragraph{Discussion about other values of $p$.} Note that Prop.~\ref{prop:conv_rsgda} and the subsequent corollaries allow for any value of $p$ in $(0, 1)$ and still guarantee the convergence of RSGDA, albeit with an additional factor in the convergence rates (See Appendix). This suggests tuning $p$ to values that are moderately larger than $1/2$, and we would still expect RSGDA to perform almost as well as with values of $p$ in the range which gives the best convergence rates. If descent steps are cheaper than ascent steps, then this strategy is sound.

\subsubsection{Interpolation setting}\label{sec:interpolation}
We consider using RSGDA in a more favorable setting, where the maximization problem is easier, as in the interpolation case, when $F$ is a finite-sum and $\tilde{\sigma}^2 = 0$.

\begin{assumption}[Interpolation]\label{asm:interpolation}
For all $(\theta, v) \in \R^d \times \R^n, \; F(\theta, v) = \frac{1}{n}\sum_{i=1}^n F_i(\theta, v)$, where for all $i \in [n]$, $F_i$ verifies Assumption~\ref{asm:smooth_str_concave} and for all $\theta \in \R^d$, there exists $v^*(\theta) \in \R^n$ such that for all $i \in [n]$, $\nabla_v F_i(\theta, v^*(\theta)) = 0$.
\end{assumption}
Note that we \textit{do not} assume that $\sigma^2 = 0$, which, by Assumption \ref{asm:noise_theta}, would have implied, for $i \in [n]$ and $(\theta, v) \in \R^d \times \R^n$ that $\nabla F_i(\theta, v) = \nabla F(\theta, v)$.

From an optimization perspective, this setting has been explored in many works on SGD \citep{vaswani2019fast, vaswani2019painless, loizou2021stochastic, sebbouh2021almost}, where it was shown that if the two previous assumptions are verified (for the single-variable objective), SGD has the same convergence rate as Gradient Descent. This setting has also recently been studied for bilinear minimax optimization \citep{li2021convergence}, 

The next result shows that in that setting, RSGDA converges at the improved rate of $\cO\br{\kappa^2\epsilon^{-4}}$ (versus $\cO\br{\kappa^3\epsilon^{-4}}$), without requiring large batch sizes.

\begin{corollary}\label{cor:interpolation}
Consider the setting of Proposition \ref{prop:conv_rsgda} and let Assumption~\ref{asm:interpolation} hold. Let $\eta_k = \eta = 1/(2L)$.
\begin{itemize}
    \item \textbf{Almost sure convergence.} Let $\zeta > 0$ and $\alpha_k = \cO\br{\frac{\eta}{\kappa^2 (k+1)^{1/2 + \zeta}}}$. Then, $\underset{t=0,\dots,k-1}{\min} \sqn{\nabla \phi(\theta_k)} =  o\br{k^{-1/2 + \zeta}}$ a.s.
    \item \textbf{Anytime convergence in expectation.} Let $\alpha_k = \cO\br{\frac{\eta}{\kappa^2 \sqrt{k+1}}}$. Then,
    $\underset{t=0,\dots,k-1}{\min} \ec{\sqn{\nabla \phi(\theta_k)}} =  \cO\br{\frac{\kappa\log(k)}{\sqrt{k+1}}}.$
    \item \textbf{Convergence in expectation for a given precision.} With a suitable choice of $\alpha$, if $p \in \left[\frac{1}{\kappa}, \frac{1}{2}\right]$ and $k \geq \Theta\br{\kappa^2\epsilon^{-4}}$, then $\underset{t=0,\dots,k-1}{\min} \ec{\norm{\nabla \phi(\theta_k)}} \leq \epsilon.$
\end{itemize}
\end{corollary}

By setting $\tilde{\sigma}^2 = 0$, all these results can be derived from Prop.~\ref{prop:conv_rsgda} in a similar fashion to the previous corollaries. As an illustration of these results, we apply our method to distributionally robust optimization in \cref{sec:dro}, where Assumption \ref{asm:interpolation} holds for the problem we consider \citep{sinha2018certifying}.

%% file: sections/applications.tex
\section{Applications} \label{sec:applications}
To illustrate our results, we  consider two nonconvex strongly concave problems: distributionally robust optimization \citep{shafieezadeh2015distributionally, kuhn2019wasserstein} and learning with a Sinkhorn loss \citep{genevay2018}.

\subsection{Distributionally robust optimization} \label{sec:dro}
The goal of distributionally robust optimization is to learn machine learning models which are robust to changes in the distribution of test data compared to training data. Consider a training dataset ${\cD \eqdef \left\{(x_1, y_1), \dots, (x_n, y_n) \right\} \subset \R^{p+q}}, p, q \geq 1$, and suppose we want to learn a robust classifier from a parametric family $\left\{f_\theta: \R^p \mapsto \R^q, \, \theta \in \R^d\right\}$. \cite{sinha2018certifying} showed that one way to do so is to solve the following optimization problem:
\begin{align}\label{eq:drw}
    \min_{\theta \in \R^d}\, \phi(\theta) &\equiv F(\theta, v), \\
    F(\theta, v) = \underset{\substack{v=[v_1,\dots,v_n]\\ \in  \R^{pn}}}{\max}\; &\frac{1}{n}\sum_{j=1}^n \ell(f_\theta(v_j), y_j) - \gamma \sqn{v_j - x_j}.
\end{align}
where $\ell: \R^p \times \R^q \mapsto \R$ is the loss function. Intuitively, the inner maximization problem requires that we find the (adversarial) inputs $\left\{v_i\right\}_i$ that maximize the loss, while minimizing the average distance to the original inputs $\left\{x_i\right\}_i$. The hyperparameter $\gamma$ controls the trade-off between these two objectives. A low $\gamma$ allows the classifier to be robust to adversarial perturbations, but sacrifices model accuracy on training data. A high $\gamma$ ensures that the model performs well on training data, but sacrifices robustness. \cite{sinha2018certifying} showed that for $\gamma$ large enough, solving this problem is nonconvex strongly concave. Moreover, we also have that Assumption~\ref{asm:interpolation} is verified.

\begin{lemma}[\cite{sinha2018certifying}]\label{lem:drw_verif_asm}
Consider $F$ defined in \eqref{eq:drw}. Assume that $(\theta, v) \mapsto \ell(f_\theta; (v_j, y_j))$ is smooth for all $(\theta, v, y) \in \R^d \times \R^{pn} \times \R^q$, and that the noise assumption \ref{asm:noise_theta} holds. Then for a large enough $\gamma$, $F$ verifies Assumption~\ref{asm:smooth_str_concave}. Moreover, since $F$ is separable in each coordinate $v_j \in \R^d, j \in [n]$, we have that Assumption~\ref{asm:interpolation} is verified. 
\end{lemma}

\begin{figure}
    \centering
    \includegraphics[scale=0.65]{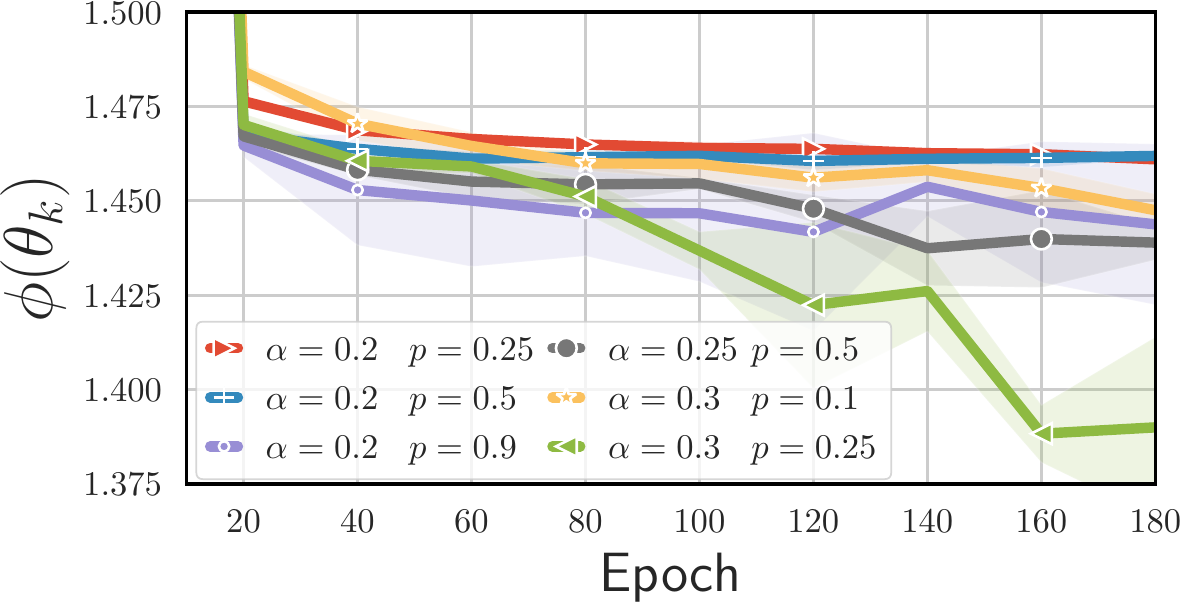}
    \caption{Training loss $\phi$ \eqref{eq:drw} when using RSGDA. $\eta = 10$. We study the effect of the descent step size $\alpha$ and the descent probability $p$. }\label{fig:mnist_perf}
    \vspace{-1.5em}
\end{figure}

\paragraph{Application to MNIST.} We reproduce the setting of \cite{sinha2018certifying} for the MNIST dataset and concentrate on the optimization aspect of their procedure. For the experiment to still be meaningful, we only considered the models which resulted in higher than 98\% validation accuracy. As done by \cite{sinha2018certifying}, we set $\gamma=1.3$ and choose $f_{\theta}$ to be a CNN with smooth ELU activations. More details on the exprimental settings can be found in the Appendix.  Contrary to \cite{lin2020}, we report the loss of interest $\phi(\theta)$ in order to better assess the effect of the parameter settings.

\begin{figure*}
    \includegraphics[width=\textwidth]{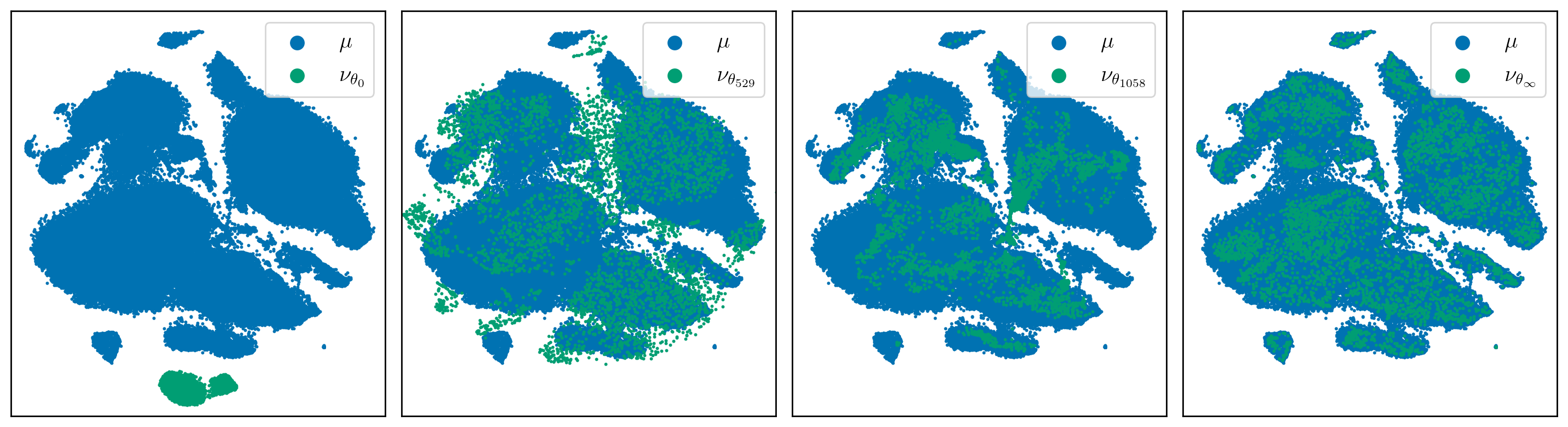}
    \caption{t-SNE embeddings of the point cloud $\nu_{\theta_k}$ \eqref{eq:sc_ot_problem} along the iterations of RSGDA ($p=0.9$).}
    \label{fig:sc_rsgda_tsne}
    \vspace{-1.5em}
\end{figure*}

\begin{figure}
    \includegraphics[scale=0.55]{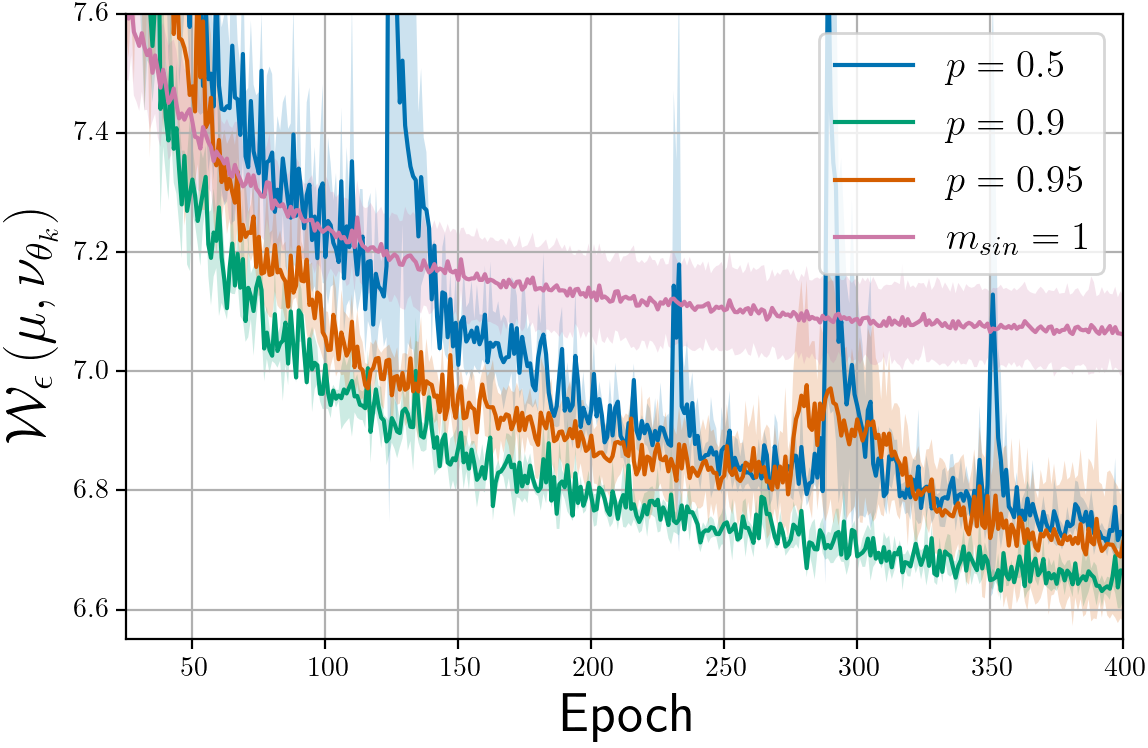}
    \caption{Loss $\cW_{\epsilon}\left(\mu, \nu_{\theta_k}\right)$ \eqref{eq:sc_ot_problem} when using RSGDA vs. Sinkhorn's algorithm, depending on the descent probability $p$ (for RSGDA) or the number of iterations $m_{\sin}$ of Sinkhorn's algorithm. $\eta=5$.}
    \label{fig:sc_rsgda_conv_plot}
    \vspace{-1.5em}
\end{figure}


Fig.~\ref{fig:mnist_perf} shows the performance of RSGDA (Alg.~\ref{alg:RSGDA}) on Problem~\eqref{eq:drw} with various parameter settings. We use minibatch RSGDA and set the ascent step size $\eta = 10$. We noticed that further tuning this hyperparameter had less impact than $\alpha$ and $p$ on the performance of RSGDA. For each probability $p$, we display the step size $\alpha$ that resulted in the lowest training loss. For all experiments with $p \geq 0.5$, choosing $\alpha > 0.25$ led the model to diverge or result in a low validation accuracy. As predicted by theory, decreasing descent probability allows to take larger step sizes ($p=0.25, \, \alpha=0.3$) and results in faster optimization, but decreasing $p$ too much ($p=0.1$) leads to a slow algorithm, as $\theta$ is updated less often. Further increasing the step size when $p=0.1$ makes RSGDA diverge.


\subsection{Learning with a Sinkhorn loss}\label{sec:applications_sinkhorn}
The second application we consider is that of learning with a semi-dual Sinkhorn loss \citep{cuturi2018semidual, kitagawa2019convergence}. We establish that this problem is indeed nonconvex-strongly concave. Then, inspired by \cite{schiebinger2019optimal, stark2020scim}, we apply semi-dual optimal transport to a single-cell data integration problem using RSGDA. But first, let us recall the definition of the regularized OT loss between two measures \citep{cuturi2013sinkhorn}.
\paragraph{Regularized OT loss.} Let $\cX$ and $\cY$ be two metric spaces. Consider two probability measures $\br{\mu, \nu} \in \cP(\cX) \times \cP(\cY)$ and let $\epsilon > 0$. The regularized OT metric between these two measures is given by
\begin{align}
    \cW_\epsilon(\mu, \nu) & \eqdef \underset{\pi \in \Pi(\mu, \nu)}{\min} \int c(x,y) \diff \pi(x,y) \\
    &+ \epsilon \int \log\br{\frac{\diff\pi(x,y)}{\diff\mu(x)\diff\nu(y)}}\diff \pi(x,y).
\end{align}
where $c: \cX \times \cY \mapsto \R_{+}$ is the cost to move a unit of mass from $x$ to $y$, $\Pi(\mu, \nu) = \left\{\pi \in \cP(\cX) \times \cP(\cY): P_{1\sharp}\pi = \mu, P_{2\sharp}\pi = \nu  \right\}$, and $P_{1}(x, y) = x$ and $P_2(x, y) = y$ are projection operators. If $\cX = \cY$, $d_\cX$ is a metric on $\cX$, and $c = {d_{\cX}}^p$, then $\cW_0^{1/p}$ defines a distance on $\cP(\cX)$. Unfortunately, computing $\cW_0$ is too costly in most applications, which justifies using the regularized OT loss with $\epsilon > 0$. Fortunately, when one of the measures is discrete, we can express $\cW_\epsilon(\mu, \nu)$ as a finite-dimensional stochastic maximization problem \citep{genevay2016}. Consider a dataset $\cY_n = \left\{y_1,\dots,y_n\right\} \cY$ and let ${\nu = \frac{1}{n}\sum_{j=1}^n \delta_{y_j}}$. Then, 
\begin{align}\label{eq:sinkhorn_prob_dual}
    \min_{\theta \in \R^d} \max_{v \in \R^n} \; &\EE{z \sim \mu}{h(z, v)}, \quad \text{where}\\
    h(x, v) \eqdef \sum_{j=1}^n \frac{v_j}{n} - \epsilon &\log\br{\sum_{j=1}^n \frac{\exp\br{\frac{v_j - c(x, y_j)}{\epsilon}}}{n}} - \epsilon.
    \vspace{-1em}
\end{align}
\textbf{Learning with a semi-discrete sinkhorn loss is a nonconvex strongly concave problem.} Consider the task of learning a parametric map from observed data $\cY_n = \left\{y_1,\dots,y_n\right\} \subset \cY$. Given a probability space $\cZ$ and a possibly continuous distribution $\mu \in \cP(\cZ)$, we may want to solve one of two problems:
\begin{gather}
    \min_{\theta \in \R^d} \cW_\epsilon\br{(f_\theta)_{\sharp}\mu, \frac{1}{n}\sum_{j=1}^n \delta_{y_j}} \equiv \max_{v \in \R^n} \; \EE{z}{h(f_\theta(z), v)},  \label{eq:sinkhorn_prob_primal}  \\
    \text{or} \quad  \min_{\theta \in \R^d} \cW_\epsilon\br{\mu, \frac{1}{n}\sum_{j=1}^n \delta_{f_\theta(y_j)}} =: \cW_{\epsilon}\br{\mu, \nu_{\theta}},\label{eq:sc_ot_problem}
\end{gather}
where $f_\theta \in \left\{f_{\theta^{\prime}} : \cZ \mapsto \cX, \; \theta^{\prime} \in \R^d\right\}$. An instance of the first problem is learning generative models \citep{genevay2018, houdard2021existence}, where one wants to fit $f_\theta$ to the dataset $\cY_n$. There, $\mu$ is typically a Gaussian in a low-dimensional space, and $\cY_n$ is for example a dataset of images. An instance of the second problem is single-cell data matching, which we present later in this section. Using the formulation~\eqref{eq:sinkhorn_prob_dual} of $\cW_\epsilon$, we can express both problems as finite dimensional stochastic min-max problems. We now show that these problems are indeed nonconvex strongly concave.
\begin{proposition}\label{lem:sinkhorn_strongly_concave}
Define for all $(\theta, v) \in \R^d \times \R^n,$ $F(\theta, v) = \EE{z \sim \mu}{h(f_\theta(z), v)}$. Assume that for all $z \in \cZ$, $\theta \mapsto c(g_z(\theta), y)$ is lipschitz continuous and smooth, and that $y \mapsto c(g_z(\theta), y)$ is lipschitz-continuous almost surely for all $(\theta, y) \in \R^d \times \cY$. 
Then, there exists $\cV \subset \R^n$ such that $F$ verifies Assumptions~\ref{asm:smooth_str_concave}, \ref{asm:noise_theta} and \ref{asm:noise_v} on $\R^d \times \cV$ and $\phi(\theta) = \max_{v \in \cV} F(\theta, v)$ for all $\theta \in \R^d$.
\end{proposition}
A similar result can be proved for~\eqref{eq:sc_ot_problem} (see App. for proof of Lem.~\ref{lem:sinkhorn_strongly_concave}). This result is of independent interest for OT practitioners since combining Lem.~\ref{lem:sinkhorn_strongly_concave} with our convergence results in \cref{sec:rsgda}, we extend~\citep{sanjabi2018convergence} to the semi-discrete setting, and we give stronger theoretical guarantees than those in \citep{houdard2021existence} when using GDA algorithms for learning with a semi-discrete Sinkhorn loss.

\textbf{Single-cell genomic data matching.} We consider the single cell melanoma tumor dataset from~\citep{stark2020scim}. Patient data is analyzed using two different technologies, scRNA and CyTOF, resulting in two point clouds of different sample sizes and dimensions. The goal is to integrate both datasets in order to understand the correspondances between technologies, and obtain a unified analysis on larger sample sizes. We denote the CyTOF point cloud as $\left\{x_1,\dots,x_m\right\} \subset \R^{41}, \, m = 135334$, and define $\mu = \frac{1}{m}\sum_{i=1}^m \delta_{x_i}$, and the scRNA one as $\left\{y_1,\dots,y_n\right\} \subset \R^{1024}, \, n = 4683$, and define $\nu = \frac{1}{n}\sum_{j=1}^n \delta_{y_j}$. Instead of embedding both point clouds in another low-dimensional space as done in~\cite{stark2020scim}, we aim to map the point cloud $\nu$ directly into $\mu$ via a multilayer perceptron (MLP), namely solve~\eqref{eq:sc_ot_problem}, where $f_\theta$ is an MLP. We compare minibatch RSGDA (where we sample from $\mu$) against Sinkhorn's algorithm (the most widely used approach to compute $\cW_\epsilon$) with minibaching \citep{cuturi2013sinkhorn} in Fig.~\ref{fig:sc_rsgda_conv_plot}. Details on the architecture and on Sinkhorn's algorithm are provided in the appendix (see also \cite{genevay2018}). Using a single iteration of Sinkhorn's algorithm resulted in the best performance for this algorithm: Due to the bias introduced by using Sinkhorn's algorithm on minibatches, increasing the number of iterations resulted in an increasingly biased gradient direction for the descent step. For RSGDA, we found that increasing the descent probability resulted in a faster algorithm up to the value $p = 0.95$, after which the algorithm is slower than $p=0.5$. On this problem, decreasing $p$ below $0.5$ was not beneficial, despite increasing the descent step size. Using t-SNE embeddings \citep{van2008visualizing}, we also display in Fig.~\ref{fig:sc_rsgda_tsne} the evolution of $\nu_{\theta_k}$ for the best run of RSGDA with $p=0.9$.



%% file: sections/conclusion.tex
\paragraph{Conclusion.} 
We have presented RSGDA, a randomized version of Epoch Stochastic Gradient Descent Ascent, a popular method for solving min-max optimization problems in machine learning. We showed empirically that RSGDA performs similarly to ESGDA, and demonstrated theoretically that RSGDA enjoys the same theoretical properties as Stochastic Gradient Descent Ascent \citep{lin2020}, a method which is well grounded in theory. We also provided practical parameter settings, which we tested numerically on a distributionally robust optimization problem and single-cell data integration using optimal transport.

%% file: sections/acknowledgements.tex
\section*{Acknowledgements}
The work of Gabriel Peyré and Othmane Sebbouh was supported in part by the French government under management of Agence Nationale de la Recherche as part of the ”Investissements d’avenir” program, reference ANR19- P3IA-0001 (PRAIRIE 3IA Institute). Gabriel Peyré also acknowledges support from the European Research Council (ERC project NORIA). Othmane Sebbouh also acknowledges the support of a ”Chaire d’excellence de l’IDEX Paris Saclay”. The authors thank Pierre Ablin for valuable feedback and discussions.

%% file: sections/appendix.tex
\appendix

\onecolumn 

\newpage

\begin{appendices}

\renewcommand{\thesection}{\Alph{section}}
\begin{center}
  \normalsize\bfseries\MakeUppercase{APPENDIX}
\end{center}

\tableofcontents

\newpage

The appendix is organized as follows:
\begin{itemize}
    \item In~\cref{sec:app_complete_results}, we present the detailed statements of the corollaries we presented in~\cref{sec:rsgda}.
    \item In~\cref{sec:app_proofs}, we present the proofs of all the statements of~\cref{sec:rsgda}.
    \item In~\cref{sec:app_semi_discrete_sinkhorn}, we present the proof of Proposition~\ref{lem:sinkhorn_strongly_concave}.
    \item In~\cref{sec:app_experiments_details}, we present details on the experimental setting of~\cref{sec:applications_sinkhorn}.
\end{itemize}

\section{Complete results}\label{sec:app_complete_results}
In this section, we give the detailed statements of the corollaries we presented in \cref{sec:rsgda}. The proofs of these results are presented in \cref{sec:app_proofs}.

\subsection{Corollary \ref{cor:conv_rgda}}
Corollary \ref{cor:conv_rgda} presented the convergence rate in expectation of Randomized Gradient Descent Ascent, \textit{i.e.} Alg.~\ref{alg:RSGDA} when using deterministic gradients to update $\theta_k$ and $v_k$.
\begin{corollary}\label{cor:conv_rgda_complete}
Consider the setting of Proposition \ref{prop:conv_rsgda}. Let $\eta_k = \eta = \frac{1}{L}$ and $\alpha = \frac{\br{1-p}}{2\kappa^2L}\frac{1}{\sqrt{p\br{2p + \frac{1-p}{\kappa}}}}$. Then,
\begin{align}
    \min_{t=0,\dots,k-1} \ec{\sqn{\nabla \phi(\theta_k)}} &\leq \frac{4\kappa^2L}{(1-p)k}\sqrt{2 + \frac{1-p}{p\kappa}} \cD_0 \\
    & + \frac{\kappa L^2}{(1-p)k}r_0. 
\end{align}
In particular, for all $p \in \left[\frac{1}{\kappa}, \frac{1}{2}\right]$, we have that 
$$\min_{t=0,\dots,k-1} \ec{\sqn{\nabla \phi(\theta_k)}} = \cO\br{\kappa^2k^{-1} }.$$
This implies that finding an $\epsilon$-stationary point requires at most $\cO(\kappa^2\epsilon^{-2})$ iterations for all $p \in \left[\frac{1}{\kappa}. \frac{1}{2}\right]$.
\end{corollary}

\subsection{Corrolary \ref{cor:exp_conv_rsgda} }
Corollary \ref{cor:exp_conv_rsgda} presented the \textit{anytime} convergence rate in expectation of RSGDA (Alg.\ref{alg:RSGDA}) using decreasing step sizes.
\begin{corollary}
Consider the setting of Proposition \ref{prop:conv_rsgda}. Then,
\begin{align}\label{eq:bound_anytime_rsgda}
    \min_{t=0,\dots,k-1} \ec{\sqn{\nabla \phi(\theta_t)}} \leq & \frac{2\cE_0}{\sum_{t=0}^{k-1}p_t \alpha_t} + \frac{4\kappa L\tilde{\sigma}^2\sum_{t=0}^{k-1}\eta_t p_t\alpha_t}{\sum_{t=0}^{k-1}p_t\alpha_t} \\
    & + \frac{2\sigma^2\kappa L}{\sum_{t=0}^{k-1} p_t \alpha_t} \sum_{t=0}^{k-1} p_t \alpha_t^2 \\
    & + \frac{2\sigma^2\kappa^4}{\sum_{t=0}^{k-1}p_t\alpha_t} \sum_{t=0}^{k-1} \br{\frac{2p_t^3  \alpha_t^3}{(1-p_t)^2\eta_t^2} + \frac{p_t^2 \alpha_t^3 \mu}{(1- p_t) \eta_t} }.
\end{align}

Let $p = p \in (0,1)$, $\eta_k = \frac{1}{2L(k+1)^{2/5}}$ and $\alpha_k = \frac{(1-p)\eta_k}{4p\kappa^2(k+1)^{1/5}}$. Then,
 $$\min_{t=0,\dots,k-1} \sqn{\nabla \phi(\theta_t)} = \cO\br{\frac{\kappa \log(k)}{k^{2/5}}}.$$
\end{corollary}

\subsection{Corollary \ref{cor:rsgda_fixed_stepsizes}}
Corollary \ref{cor:rsgda_fixed_stepsizes} presented the convergence rate in expectation of RSGDA (Alg.\ref{alg:RSGDA}) given a specified precision $\epsilon > 0$ for an arbitrary minibatch size.
\begin{corollary}
Consider the setting of Proposition \ref{prop:conv_rsgda}. Choose following step sizes
\begin{align}
    \eta = \frac{\epsilon^2}{24\kappa L \sigma^2}\quad \text{and} \quad \alpha = \frac{\epsilon^2}{12\kappa L \sigma^2} \min\left\{
    \frac{\epsilon^2}{12\kappa L \sigma^2}, \, \frac{1-p}{p}\frac{\epsilon^3}{48\sqrt{3}\kappa^3L\sigma^3}, \, \sqrt{\frac{1-p}{p}}\frac{\epsilon^2}{12\sqrt{2}\kappa^2 L \sigma^2}, \, \frac{1-p}{8\kappa^2L\sqrt{p\br{2p + \frac{(1-p)\epsilon^2}{24\kappa^2\sigma^2}}}}
    \right\},
\end{align}
and a number of iterations $k$ which verifies
\begin{align}
    k \geq \frac{12}{\epsilon^2} \left\{ \frac{12\cD_0\kappa L \sigma^2}{p\epsilon^2}, \, \frac{48\sqrt{3}\cD_0\kappa^3L\sigma^3}{(1-p)\epsilon^3}, \, \frac{12\sqrt{2}\cD_0\kappa^2 L \sigma^2 }{\sqrt{p(1-p)}\epsilon^2}, \frac{8\kappa^2L\sqrt{2 + \frac{(1-p)\epsilon^2}{24p\kappa^2\sigma^2}}}{1-p}, \, \frac{\kappa^2L^2r_0\sigma^2}{(1-p)\sigma^2} \right\}.
\end{align}
Then, by choosing $p \in \left[\frac{\epsilon}{\kappa^2}, \frac{1}{2}\right]$, we have that finding an $\epsilon$-stationary point requires at most $\cO\br{\kappa^3\epsilon^{-5}}$ iterations.
\end{corollary}

\subsection{Corollary \ref{cor:rsgda_big_batch}}
Corollary \ref{cor:rsgda_big_batch} presented the convergence rate in expectation of RSGDA (Alg.\ref{alg:RSGDA}) given a specified precision $\epsilon > 0$ for a large enough minibatch size.
\begin{corollary}[Large minibatch sizes]\label{cor:app_rsgda_big_batch}
Consider the setting of proposition \ref{prop:conv_rsgda}. Let Assumption \ref{asm:noise_large_batch} hold. Choose the step sizes
\begin{align}
    \eta = \frac{1}{L} \quad \mbox{and} \quad \alpha = \frac{1-p}{2\kappa^2L\sqrt{p\br{2p + \frac{1-p}{\kappa}}}},
\end{align}
There exists a minibatch size $M(\kappa, \epsilon, p)$ such that if the total number of stochastic gradient evaluations is larger than
\begin{align}
    \cO\br{ \frac{\frac{\kappa^2 L \cD_0 }{1-p}\sqrt{2 + \frac{1-p}{p\kappa}} + \frac{\kappa L^2 r_0}{1-p}}{\epsilon^2}\max\left\{1, M(\kappa, \epsilon, p)\right\}},
\end{align}
then, $\min_{t=0,\dots, k-1} \ec{\norm{\nabla \phi(\theta_t)}} \leq \epsilon$.
Moreover, choosing $p \in \left[\frac{1}{\kappa},\frac{1}{2}\right]$ ensures that the total number stochastic gradient computations is $\cO\br{\kappa^3\epsilon^{-4}}$.
\end{corollary}

\section{Proofs}\label{sec:app_proofs}

\subsection{Proof of Proposition \ref{prop:conv_rsgda}}
\begin{proof}
In this proof, we define $\theta_k^+ = \theta_k - \alpha_k \nabla_\theta f(\theta_k, v_k, z_k)$ and $v_k^+ = \Pi_\cV \br{v_k + \eta_k \nabla_v f(\theta_k, v_k; z_k)}$. We denote by $\EE{z_k}{\cdot}$ the expectation conditioned on the random variable $z_k$, and by $\EE{k}{\cdot}$ the expectation conditioned on all past random variables.

Recall that
\begin{eqnarray*} (\theta_{k+1}, v_{k+1})
    =\left\{
    \begin{array}{ll}
        (\theta_k^+, \; v_k) & \mbox{\textbf{w. p.} }p \\
        (\theta_k \;, v_k^+) & \mbox{\textbf{w. p.} } 1-p
    \end{array} \right.
    \end{eqnarray*}

We have
\begin{align}\label{eq:phi_theta_k+1_p}
    \EE{k}{\phi(\theta_{k+1})} = p \EE{z_k}{\phi(\theta_k^+)} + (1 - p) \phi(\theta_k).
\end{align}

From Lemma \ref{lem:general_smooth_lip}, we have that $\phi$ is $2\kappa L$-smooth. Hence,
\begin{align}
    \phi(\theta_{k}^+) &\leq \phi(\theta_k) - \alpha_k \langle \nabla \phi(\theta_k), \nabla_\theta f(\theta_k, v_k; z_k) \rangle +  \alpha_k^2 \kappa L \sqn{\nabla_\theta f(\theta_k, v_k; z_k)}.
\end{align}
Thus,
\begin{align}
    \EE{z_k}{\phi(\theta_{k}^+)} &\leq \phi(\theta_k) - \alpha_k \langle \nabla \phi(\theta_k), \nabla_\theta F(\theta_k, v_k) \rangle + \alpha_k^2\kappa L \EE{z_k}{\sqn{\nabla_\theta f(\theta_k, v_k; z_k)}}\\
    &= \phi(\theta_k) - \frac{\alpha_k}{2} \sqn{\nabla \phi(\theta_k)} - \frac{\alpha_k}{2}\sqn{\nabla_\theta F(\theta_k, v_k)} + \frac{\alpha_k}{2} \sqn{\nabla \phi(\theta_k) - \nabla_\theta F(\theta_k, v_k)} \\
    & + \alpha_k^2\kappa L \EE{z_k}{\sqn{\nabla_\theta f(\theta_k, v_k; z_k) - \nabla_\theta F(\theta_k, v_k)}} +\alpha_k^2\kappa L \sqn{\nabla_\theta F(\theta_k, v_k)}\\
    &\leq \phi(\theta_k) - \frac{\alpha_k}{2}\sqn{\nabla \phi(\theta_k)} + \frac{\alpha_k}{2}\sqn{\nabla \phi(\theta_k) -  \nabla_\theta F(\theta_k, v_k)} \\
    &- \frac{\alpha_k}{2}\br{1 - 2\alpha_k\kappa L}\sqn{\nabla_\theta F(\theta_k, v_k)} + \alpha_k^2\kappa L \sigma^2 \\
    &\leq \phi(\theta_k) - \frac{\alpha_k}{2}\sqn{\nabla \phi(\theta_k)} + \frac{\alpha_k}{2}\sqn{\nabla \phi(\theta_k) -  \nabla_\theta F(\theta_k, v_k)} + \alpha_k^2\kappa L \sigma^2  \\
    &- \frac{\alpha_k}{4}\sqn{\nabla_\theta F(\theta_k, v_k)}, \label{eq:theta_step1}
\end{align}
where we used the fact that $\alpha_k \leq \frac{1}{4\kappa L}$ in the last inequality.

Let $\delta_k \eqdef \sqn{v_k - v^*(\theta_k)}$. From assumption \ref{asm:smooth_str_concave}, $F$ is $L$-smooth. Hence,
\begin{align}
    \EE{z_k}{\phi(\theta_k^+)} &\leq \phi(\theta_k) - \frac{\alpha_k}{2}\sqn{\nabla \phi(\theta_k)} + \frac{\alpha_kL^2}{2}\delta_k + \alpha_k^2\kappa L \sigma^2  - \frac{\alpha_k}{4}\sqn{\nabla_\theta F(\theta_k, v_k)}.
\end{align}
Using this inequality in \eqref{eq:phi_theta_k+1_p}, we have
\begin{align}
    \EE{k}{\phi(\theta_{k+1}} \leq \phi(\theta_k) - \frac{p\alpha_k}{2}\sqn{\nabla \phi(\theta_k)} + \frac{p \alpha_kL^2}{2}\delta_k + p\alpha_k^2\kappa L \sigma^2  - \frac{p\alpha_k}{4}\sqn{\nabla_\theta F(\theta_k, v_k)}.
\end{align}\label{eq:theta_step2}

Besides, we have
\begin{align}
    \EE{k}{\delta_{k+1}} = (1 - p) \EE{z_k}{\sqn{v_k^+ - v^*(\theta_k)}} + p \EE{z_k}{\sqn{v_k - v^*(\theta_{k}^+)}}.
\end{align}
First, note that since $v_k^{+} = \Pi_\cV \br{v_k + \eta_k \nabla_v f(\theta_k, v_k; z_k)}$ and $v^*(\theta_k) \in \cV$, where $\cV$ is a convex set, we have that $$\sqn{v_k^+ - v^*(\theta_k)} = \sqn{\Pi_\cV \br{v_k + \eta_k \nabla_v f(\theta_k, v_k; z_k)} - \Pi_\cV\br{v^*(\theta_k)}} \leq \sqn{v_k + \eta_k \nabla_v f(\theta_k, v_k; z_k)  - v^*(\theta_k)},$$
where we used the fact that $\Pi_\cV$ is contractive. Hence, using the classical analysis for SGD in the strongly convex and smooth setting (see for example \cite{gower2019sgd}), we have that
\begin{align}
    \EE{z_k}{\sqn{v_k^+ - v^*(\theta_k)}} \leq (1 - \eta_k\mu)\sqn{v_k - v^*(\theta_k)} - 2\eta_k\br{1 - 2\eta_k L}\br{f(\theta_k, v^*(\theta_k)) -  f(\theta_k, v_k)} + 2\eta_k^2 \tilde{\sigma}^2.
\end{align}
Hence, with $\eta_k \leq \frac{1}{2L}$, we have
\begin{align}
    \EE{k}{\delta_{k+1}} \leq (1 - p)(1 - \eta_k\mu)\sqn{v_k - v^*(\theta_k)} + p \EE{z_k}{\sqn{v_k - v^*(\theta_{k}^+)}} + 2(1 - p)\eta_k^2 \tilde{\sigma}^2.
\end{align}
Let $\beta_k > 0$. Then,
\begin{align}
    \EE{k}{\delta_{k+1}} &\leq \br{(1 - p)(1 - \eta_k\mu) + p\br{1+\beta_k}} \delta_k+ p\br{1+\frac{1}{\beta_k}} \EE{z_k}{\sqn{v^*(\theta_{k}^+) - v^*(\theta_k)}} + 2(1 - p)\eta_k^2 \tilde{\sigma}^2.
\end{align}
From Lemma \ref{lem:general_smooth_lip}, we have that $v^*$ is $\kappa$-lipschitz. Hence, by also using Assumption \ref{asm:noise_theta},
\begin{align}
    \EE{z_k}{\sqn{v^*(\theta_{k}^+) - v^*(\theta_k)}} \leq \alpha_k^2 \kappa^2 \EE{z_k}{\sqn{\nabla_\theta f(\theta_k, v_k; z_k)}} \leq \alpha_k^2\kappa^2 \sqn{\nabla_\theta F(\theta_k, v_k)} + \alpha_k^2\kappa^2\tilde{\sigma}^2.
\end{align}
Hence,
\begin{align}
    \EE{k}{\delta_{k+1}} &\leq \br{(1 - p)(1 - \eta_k\mu) + p\br{1+\beta_k}} \delta_k+ p\br{1+\frac{1}{\beta_k}}\alpha_k^2 \kappa^2 \sqn{\nabla_\theta F(\theta_k, v_k)} \\
    & + 2 (1 - p)\eta_k^2 \tilde{\sigma}^2 + p\br{1+\frac{1}{\beta_k}}\alpha_k^2 \kappa^2  \sigma^2.
\end{align}
With $\beta_k = \frac{\br{1- p}\eta_k \mu}{2p}$, this inequality becomes
\begin{align}
    \EE{k}{\delta_{k+1}} &\leq \br{1 - \frac{\br{1-p}\eta_k\mu}{2}} \delta_k+ \frac{p\br{2p + \br{1-p}\eta_k\mu}\alpha_k^2\kappa^2}{(1-p)\eta_k\mu} \sqn{\nabla_\theta F(\theta_k, v_k)} \\
    & + 2(1 - p)\eta_k^2 \tilde{\sigma}^2 + \frac{p\br{2p + \br{1-p}\eta_k\mu}\alpha_k^2\kappa^2}{(1-p)\eta_k\mu}   \sigma^2.
\end{align}
Rearranging, we have 
\begin{align}
    \delta_k &\leq \br{\frac{2}{(1-p)\eta_k\mu}\delta_k - \frac{2}{(1-p)\eta_k\mu}\EE{k}{\delta_{k+1}}} + \frac{2p\br{2p + \br{1-p}\eta_k\mu}\alpha_k^2\kappa^2}{(1-p)^2\eta_k^2\mu^2} \sqn{\nabla_\theta F(\theta_k, v_k)} \\
    & + \frac{4\eta_k \tilde{\sigma}^2}{\mu} + \frac{2p\br{2p + \br{1-p}\eta_k\mu}\alpha_k^2\kappa^2\sigma^2}{(1-p)^2\eta_k^2\mu^2}.
\end{align}
Hence,
\begin{align}
    \frac{p\alpha_kL^2}{2}\delta_k &\leq \br{\frac{p\alpha_k\kappa L}{(1-p)\eta_k}\delta_k - \frac{p\alpha_k\kappa L}{(1-p)\eta_k}\EE{k}{\delta_{k+1}}} + \frac{p^2\br{2p + \br{1-p}\eta_k\mu}\alpha_k^3\kappa^4}{(1-p)^2\eta_k^2} \sqn{\nabla_\theta F(\theta_k, v_k)} \\
    & + 2\eta_k p \alpha_k \kappa L\tilde{\sigma}^2 + \frac{p^2\br{2p + \br{1-p}\eta_k\mu}\alpha_k^3\kappa^4 \sigma^2}{(1-p)^2\eta_k^2}\\
    &\leq \br{\frac{p\alpha_k\kappa L}{(1-p)\eta_k}\delta_k - \frac{p\alpha_{k+1}\kappa L}{(1-p)\eta_{k+1}}\EE{k}{\delta_{k+1}}} + \frac{p^2\br{2p + \br{1-p}\eta_k\mu}\alpha_k^3\kappa^4}{(1-p)^2\eta_k^2} \sqn{\nabla_\theta F(\theta_k, v_k)} \\
    & + 2\eta_k p \alpha_k \kappa L\tilde{\sigma}^2 + \frac{p^2\br{2p + \br{1-p}\eta_k\mu}\alpha_k^3\kappa^4 \sigma^2}{(1-p)^2\eta_k^2},
\end{align}
where we used in the last inequality that $\frac{\alpha_{k+1}}{\eta_{k+1}} \leq \frac{\alpha_{k}}{\eta_{k}}$
Using this inequality in \eqref{eq:theta_step2} and rearranging gives

\begin{align}
    &\frac{p\alpha_k}{2}\sqn{\nabla \phi(\theta_k)} + \EE{k}{\phi(\theta_{k+1}) + \frac{p\alpha_{k+1}\kappa L}{(1-p)\eta_{k+1}}\delta_{k+1}} \\
    &\leq \phi(\theta_k) + \frac{p\alpha_k\kappa L}{(1-p)\eta_k}\delta_k - \frac{\alpha_k p}{4}\br{1 - \frac{4p\br{2p + \br{1-p}\eta_k\mu}\alpha_k^2\kappa^4}{(1-p)^2\eta_k^2}}\sqn{\nabla_\theta F(\theta_k, v_k)}\\
    & + p\alpha_k^2\kappa L\sigma^2 + 2\eta_k p \alpha_k \kappa L\tilde{\sigma}^2 + \frac{p^2\br{2p + \br{1-p}\eta_k\mu}\alpha_k^3\kappa^4 \sigma^2}{(1-p)^2\eta_k^2}.
\end{align}
Hence, since $\alpha_k \leq \frac{(1-p)\eta_k}{4\kappa^2\sqrt{p\br{2p + (1-p)\eta_k\mu}}}$,
\begin{align}
    &p\alpha_k\sqn{\nabla \phi(\theta_k)} + \EE{k}{2\phi(\theta_{k+1}) + \frac{2p\alpha_{k+1}\kappa L}{(1-p)\eta_{k+1}}\delta_{k+1}} \\
    &\leq 2\phi(\theta_k) + \frac{2 p\alpha_k\kappa L}{(1-p)\eta_k}\delta_k \\
    & + 2p\alpha_k^2\kappa L\sigma^2 + 4\eta_k p \alpha_k \kappa L\tilde{\sigma}^2 + \frac{2p^2\br{2p + \br{1-p}\eta_k\mu}\alpha_k^3\kappa^4 \sigma^2}{(1-p)^2\eta_k^2}.
\end{align}
Hence,
\begin{align}
    p\alpha_k \sqn{\nabla \phi(\theta_k)} + 2\EE{k}{\cE_{k+1}} \leq 2\cE_{k} + 2\sigma^2\br{ p\alpha_k^2\kappa L + \frac{p^2\br{2p + \br{1-p}\eta_k\mu}\alpha_k^3\kappa^4}{(1-p)^2\eta_k^2}} + 4\eta_k p \alpha_k \kappa L\tilde{\sigma}^2.
\end{align}

\end{proof}

\subsection{Proof of Corollary \ref{cor:as_conv_rsgda}}

Before the proof of Corollary~\ref{cor:as_conv_rsgda}, we first present a simplified version of the Robbins-Siegmund theorem.
\begin{lemma}[\cite{Robbins71}]\label{lem:simple_RS}
Consider a filtration $\br{\cF_k}_k$, the  nonnegative sequences of $\br{\cF_k}_k-$adapted processes $\br{V_k}_k$,  $\br{U_k}_k$ and $\br{Z_k}_k$ such that $\sum_k Z_k < \infty \; \mbox{almost surely}$, and \[\forall k \in \N, \; \ec{V_{k+1}|\cF_k} + U_{k+1} \leq  V_k + Z_k.\]
Then $\br{V_k}_k$ converges and $\sum_k U_k < \infty $ \textit{almost surely}.
\end{lemma}

We now move on to the proof of Corollary~\ref{cor:as_conv_rsgda}.
\begin{proof}
In this proof, we use a similar proof technique to \cite{sebbouh2021almost}. From Proposition \ref{prop:conv_rsgda}, we have 
\begin{align}
    & \alpha_k \sqn{\nabla \phi(\theta_k)} + \frac{2\EE{k}{\cE_{k+1}}}{p} \leq \frac{2\cE_{k}}{p} + 4\eta_k \alpha_k \kappa L\tilde{\sigma}^2  + 2\sigma^2\br{\alpha_k^2\kappa L + \frac{p\br{2p + \br{1-p}\eta_k\mu}\alpha_k^3\kappa^4}{(1-p)^2\eta_k^2}}.
\end{align}
Using Lemma \ref{lem:simple_RS} and the fact that $\sum_k \alpha_k^2\sigma^2 < \infty$, $\sum_k \eta_k \alpha_k\tilde{\sigma}^2 < \infty$,  $\sum_k \frac{\alpha_k^3}{\eta_k^2}\sigma^2 < \infty$, $\sum_k \frac{\alpha_k^3}{\eta_k}\sigma^2 < \infty$, we have that $\br{\cE_k}_k$ converges \textit{almost surely}. Now define for all $k \in \N$,
\begin{align}
    w_k = \frac{2\alpha_k}{\sum_{j=0}^{k}\alpha_j}, \quad g_0 = \sqn{\nabla \phi(\theta_0)}, \quad g_{k+1} = (1-w_k)g_k + w_k\sqn{\nabla \phi(\theta_k)}.
\end{align}
Notice that since $\br{\alpha_k}_k$ is decreasing, we have $w_k \in [0, 1]$. Hence, using the convexity of the squared norm, we have
\begin{align}
     \frac{\sum_{j=0}^k \alpha_j}{2}g_{k+1} + \frac{2\EE{k}{\cE_{k+1}}}{p} + \frac{\alpha_k}{2}g_k  &\leq \frac{\sum_{j=0}^{k-1} \alpha_j}{2}g_k + \frac{2\cE_{k}}{p} \\
    &+ 4\eta_k \alpha_k \kappa L\tilde{\sigma}^2  + 2\sigma^2\br{\alpha_k^2\kappa L + \frac{p\br{2p + \br{1-p}\eta_k\mu}\alpha_k^3\kappa^4}{(1-p)^2\eta_k^2}}.
\end{align}
Using Lemma \ref{lem:simple_RS} and the step size conditions again, and the fact that $\br{\cE_k}_k$ converges \textit{almost surely}, gives that $\br{\sum_{j=0}^k \alpha_jg_{k+1}}_k$ converges \textit{almost surely} and that $\sum_k \alpha_k g_k < \infty$ \textit{almost surely}. In particular, this implies that $\lim_k \alpha_k g_k = 0$. Notice that $\alpha_k g_k = \frac{\alpha_k}{\sum_{j=0}^{k-1}\alpha_j} \sum_{j=0}^{k-1}\alpha_j g_k$. Hence, since we have that $\br{\sum_{j=0}^k \alpha_jg_{k+1}}_k$ converges \textit{almost surely} and $\sum_k \frac{\alpha_k}{\sum_{j=0}^{k-1}\alpha_j} = \infty$ (which is a consequence of the fact that $\sum_k \alpha_k = \infty$), then $\lim_k \sum_{j=0}^{k-1} \alpha_j g_k = 0$, \textit{i.e.} $$g_k = o\br{\frac{1}{\sum_{j=0}^{k-1}\alpha_j}}$$
Finally, since, $g_k$ is a weighted average of $\left\{\sqn{\nabla \phi(\theta_0)}, \dots, \sqn{\nabla \phi(\theta_{k-1})}\right\}$, we have that $g_k \geq \min_{t=0,\dots, k-1} \sqn{\nabla \phi(\theta_t)}$. Hence, $$\min_{t=0,\dots, k-1} \sqn{\nabla \phi(\theta_t)} = o\br{\frac{1}{\sum_{j=0}^{k-1}\alpha_j}}.$$

\end{proof}

\subsection{Proofs for convergence rates in expectation}
All the convergence in expectation proofs follow by telescopic cancellation in \eqref{eq:conv_reccurrence_rsgda}. Indeed, summing Inequality \eqref{eq:conv_reccurrence_rsgda} between $t=0$ and $k-1$, and using the fact that $\forall t=0, \dots, k-1, \; \sqn{\nabla \phi(\theta_t)} \geq \min_{j=0,\dots,k-1} \sqn{\nabla \phi(\theta_j)}$, we have
\begin{align}
    \min_{t=0,\dots,k-1} \sqn{\nabla \phi(\theta_t)} \leq & \frac{2\br{\phi(\theta_0) - \phi_*}}{p\sum_{t=0}^{k-1} \alpha_t} + \frac{2\alpha_0\kappa L\sqn{v_0 - v^*(\theta_0)}}{(1-p)\eta_0\sum_{t=0}^{k-1} \alpha_t} \\
    & + \frac{4\kappa L\tilde{\sigma}^2\sum_{t=0}^{k-1}\eta_k \alpha_t}{\sum_{t=0}^{k-1}\alpha_t} + \frac{2\kappa L\sigma^2\sum_{t=0}^{k-1}\alpha_t^2}{\sum_{t=0}^{k-1}\alpha_t} \\
    & + \frac{2\sigma^2\kappa^4 p}{(1-p)^2\sum_{t=0}^{k-1}\alpha_t} \br{2p\sum_{t=0}^{k-1}\frac{\alpha_t^3}{\eta_t^2} + (1-p)\mu\sum_{t=0}^{k-1}\frac{\alpha_t^3}{\eta_t}}. \label{eq:inequality_after_telescoping}
\end{align}

\subsubsection{Proof of Corollary \ref{cor:conv_rgda}}
Using constant step sizes in \eqref{eq:inequality_after_telescoping}, we have
\begin{align}\label{eq:inequality_after_telescoping_fixed_step_sizes}
    \min_{t=0,\dots,k-1} \ecn{\nabla \phi(\theta_t)} \leq \frac{2\cD_0}{\alpha p k} + \frac{2\kappa L r_0}{(1-p)\eta k} + 4\eta \kappa L \tilde{\sigma}^2 + 2\alpha\kappa L \sigma^2 + \frac{2\alpha^2\kappa^4p^2\sigma^2}{(1-p)^2\eta^2} + \frac{2\alpha^2\kappa^3 Lp\sigma^2}{(1-p)\eta}.
\end{align}

\begin{proof}
When using the exact gradients, we have that $\sigma^2 = \tilde{\sigma}^2 = 0$. Using the constant step sizes of Corollary \ref{cor:conv_rgda} in \eqref{eq:inequality_after_telescoping_fixed_step_sizes} directly gives the desired result.
\end{proof}

\subsubsection{Proof of Corollary \ref{cor:exp_conv_rsgda}}
\begin{proof}
Using Inequality \eqref{eq:inequality_after_telescoping} and the parameter settings of Corollary \ref{cor:exp_conv_rsgda}, we have
\begin{align}
    \min_{t=0,\dots,k-1} \ec{\sqn{\nabla \phi(\theta_t)}} \lessapprox & \frac{2\br{\phi(\theta_0) - \phi_*}}{p\alpha_0 (k+1)^{2/5}} + \frac{2\kappa L\sqn{v_0 - v^*(\theta_0)}}{(1-p)\eta_0 (k+1)^{2/5}} \\
    & + \frac{4 \kappa L \tilde{\sigma}^2\eta_0\log(k+1)}{(k+1)^{2/5}} + \frac{2\kappa L \sigma^2 \alpha_0}{(k+1)^{2/5}} \\
    & + \frac{2\sigma^2\kappa^4 p\alpha_0^2}{\eta_0(1-p)^2
    (k+1)^{2/5}} \br{\frac{2p\log(k+1)}{\eta_0} + (1-p)\mu},
\end{align}
where $"\lessapprox"$ indicates that we omit the absolute constants arising in the summations. The asymptotically dominant term is $\frac{4 \kappa L \tilde{\sigma}^2\eta_0\log(k+1)}{(k+1)^{2/5}}$. Hence
\begin{align}
     \min_{t=0,\dots,k-1} \ec{\sqn{\nabla \phi(\theta_t)}} = \cO\br{\frac{\kappa\log(k)}{\sqrt{k}}}
\end{align}
\end{proof}

\subsubsection{Proof of Corollary \ref{cor:rsgda_fixed_stepsizes}} \label{sec:app_proof_cor_rsgda_fixed_step_sizes}
\begin{proof}
For simplicity, we consider that $\sigma^2 \geq \tilde{\sigma}^2 \geq 1$. Otherwise, we can simply replace $\sigma$ and $\tilde{\sigma}$ by $\max\left\{\sigma, \tilde{\sigma}, 1\right\}$, and the proof will still hold with $\max\left\{\sigma, \tilde{\sigma}, 1\right\}$ instead of $\sigma$.

Let $\epsilon > 0$. The proof follows simply from forcing each term of the LHS of \eqref{eq:inequality_after_telescoping_fixed_step_sizes} to be smaller than $\frac{\epsilon^2}{6}$.
This results in the following step sizes
\begin{align}
    \eta = \frac{\epsilon^2}{24\kappa L \sigma^2}\quad \text{and} \quad \alpha = \frac{\epsilon^2}{12\kappa L \sigma^2} \min\left\{
    \frac{\epsilon^2}{12\kappa L \sigma^2}, \, \frac{1-p}{p}\frac{\epsilon^3}{48\sqrt{3}\kappa^3L\sigma^3}, \, \sqrt{\frac{1-p}{p}}\frac{\epsilon^2}{12\sqrt{2}\kappa^2 L \sigma^2}, \, \frac{1-p}{8\kappa^2L\sqrt{p\br{2p + \frac{(1-p)\epsilon^2}{24\kappa^2\sigma^2}}}}
    \right\},
\end{align}
and the following lower bound on $k$
\begin{align}
    k \geq \frac{12}{\epsilon^2} \left\{ \frac{12\cD_0\kappa L \sigma^2}{p\epsilon^2}, \, \frac{48\sqrt{3}\cD_0\kappa^3L\sigma^3}{(1-p)\epsilon^3}, \, \frac{12\sqrt{2}\cD_0\kappa^2 L \sigma^2 }{\sqrt{p(1-p)}\epsilon^2}, \frac{8\kappa^2L\sqrt{2 + \frac{(1-p)\epsilon^2}{24p\kappa^2\sigma^2}}}{1-p}, \, \frac{\kappa^2L^2r_0\sigma^2}{(1-p)\sigma^2} \right\}.
\end{align}
By choosing $p \in \left[\frac{\epsilon}{\kappa^2}, \frac{1}{2}\right]$, we have that the RHS is at most of the order $\Theta\br{\kappa^3\epsilon^{-5}}$.
\end{proof}

\subsubsection{Proof of Corollary \ref{cor:rsgda_big_batch}}
To prove Corollary \ref{cor:rsgda_big_batch}, we need to use the following lemma, which ensures that the variance of the stochastic gradients decreases linearly with the minibatch size.
\begin{lemma}[Lemma A.2 in \cite{lin2020}]
Let Assumption \ref{asm:noise_large_batch} hold. Then, is $G_z(\theta, v) = \frac{1}{M}\sum_{i=1}^M \nabla_\theta f(\theta, v; z^i)$ where $z_1,\dots,z_M$ are sampled i.i.d, then
\begin{align}
    \ec{\sqn{G_z(\theta, v) - \nabla F(\theta, v)}} \leq \frac{\bar{\sigma}^2}{M}.
\end{align}
And the same holds with the gradient with respect to $v$.
\end{lemma}
\begin{proof}
Using \eqref{eq:inequality_after_telescoping_fixed_step_sizes} with the constant step sizes of Corollary \ref{cor:rsgda_big_batch} and Lemma
\begin{align}
    \min_{t=0,\dots,k-1}\ecn{\nabla \phi(\theta_t)} &\leq \frac{8\kappa^2L}{(1-p)k}\sqrt{2 + \frac{1-p}{2p\kappa}}\cD_0 + \frac{2\kappa L^2}{(1-p)k}r_0 \\
    &+ \frac{2\sigma^2}{M}\br{\frac{1-p}{2\kappa\sqrt{p\br{2p + \frac{1-p}{2\kappa}}}} + 2\kappa + \frac{p}{2\kappa^2\br{2p + \frac{1-p}{2\kappa}}} + \frac{1-p}{4\kappa\br{2p + \frac{1-p}{2\kappa}}}}.
\end{align}
Hence, since $\frac{p}{2\kappa^2\br{2p + \frac{1-p}{2\kappa}}} + \frac{1-p}{4\kappa\br{2p + \frac{1-p}{2\kappa}}} \leq 2\kappa$, we have
\begin{align}
    \min_{t=0,\dots,k-1}\ecn{\nabla \phi(\theta_t)} &\leq \frac{8\kappa^2L}{(1-p)k}\sqrt{2 + \frac{1-p}{2p\kappa}}\cD_0 + \frac{2\kappa L^2}{(1-p)k}r_0 \\
    &+ \frac{2\kappa\sigma^2}{M}\br{\frac{1-p}{2\kappa^2\sqrt{p\br{2p + \frac{1-p}{2\kappa}}}} + 4}.
\end{align}
Choosing $k \geq \frac{2}{\epsilon^2}\br{ \frac{_\kappa^2L}{(1-p)k}\sqrt{2 + \frac{1-p}{2p\kappa}}\cD_0 + \frac{2\kappa L^2}{(1-p)k}r_0}$ and $M = M\br{\kappa, \epsilon}$ ensures that 
\begin{align}
  \frac{8\kappa^2L}{(1-p)k}\sqrt{2 + \frac{1-p}{2p\kappa}}\cD_0 + \frac{2\kappa L^2}{(1-p)k}r_0 \leq \frac{\epsilon^2}{2} \quad \mbox{and} \quad \frac{2\kappa\sigma^2}{M}\br{\frac{1-p}{2\kappa^2\sqrt{p\br{2p + \frac{1-p}{2\kappa}}}} + 4} \leq \frac{\epsilon^2}{2}.
\end{align}
Hence, the total number of samples required to guarantee that $\min_{t=0,\dots,k-1}\ec{\norm{\nabla \phi(\theta_t)}} \leq \epsilon$ is 

\begin{align}
    \max\left\{1, M(\kappa, \epsilon)\right\} k  = \cO\br{ \frac{\frac{\kappa^2 L \cD_0 }{1-p}\sqrt{2 + \frac{1-p}{2p\kappa}} + \frac{\kappa L^2 r_0}{1-p}}{\epsilon^2}\max\left\{1, M(\kappa, \epsilon)\right\}}.
\end{align}

\end{proof}

\subsection{Proofs for the interpolation results of Section \ref{sec:interpolation} (Corollary \ref{cor:interpolation})}

\subsubsection{Almost sure convergence}
\begin{proof}
Using $\tilde{\sigma}^2 = 0$ and $\eta_k = \frac{1}{2L}$ in \eqref{eq:conv_reccurrence_rsgda}, we have that 
\begin{align}
    p\alpha_k \sqn{\nabla \phi(\theta_k)} + 2\EE{k}{\cE_{k+1}} \leq 2\cE_{k} + 2\alpha_k^2\sigma^2\br{ p\kappa L + \frac{4p^2\br{2p + \frac{\br{1-p}}{\kappa}}\alpha_k\kappa^4L^2}{(1-p)^2}}.
\end{align}
Now note that by choosing $\alpha_k = \frac{\eta}{4(k+1)^{1/2 - \epsilon}\sqrt{p\br{2p + (1-p)\eta \mu}}}$, we have that
\begin{align}
    \sum_k \alpha_k = \infty, \quad \sum_{k} \alpha_k^2 < \infty \quad \text{and} \quad  \sum_{k} \alpha_k^3 < \infty.
\end{align}
Thus, proceeding as in the proof of Corollary \ref{cor:as_conv_rsgda}, but with different choices of step sizes, we have that
\begin{align}
    \min_{t=0,\dots,k-1} \sqn{\nabla \phi(\theta_k)} = o\br{k^{-1/2 + \zeta}}.
\end{align}
\end{proof}

\subsubsection{Anytime convergence in expectation}
\begin{proof}
Using $\tilde{\sigma}^2 = 0$ and $\eta_k = \frac{1}{2L}$ in \eqref{eq:conv_reccurrence_rsgda}, we have that
\begin{align}
    \min_{t=0,\dots,k-1} \ec{\sqn{\nabla \phi(\theta_t)}} \leq & \frac{2\br{\phi(\theta_0) - \phi_*}}{p\sum_{t=0}^{k-1} \alpha_t} + \frac{4\alpha_0\kappa L^2\sqn{v_0 - v^*(\theta_0)}}{(1-p)\sum_{t=0}^{k-1} \alpha_t} \\
    & + \frac{2\kappa L\sigma^2\sum_{t=0}^{k-1}\alpha_t^2}{\sum_{t=0}^{k-1}\alpha_t} + \frac{2\sigma^2\kappa^4 p}{(1-p)^2\sum_{t=0}^{k-1}\alpha_t} \br{8pL^2\sum_{t=0}^{k-1}\alpha_t^3 + 2(1-p)L\mu\sum_{t=0}^{k-1}\alpha_t^3}.
\end{align}
Hence, choosing $\alpha_k = \frac{\eta}{4(k+1)^{1/2}\sqrt{p\br{2p + (1-p)\eta \mu}}}$, we have, omitting absolute constants. As is the case in the proof of Corollary \ref{cor:exp_conv_rsgda}, the asymptotically dominant term in the RHS is $\frac{\sum_{t=0}^{k-1}\alpha_t^2}{\sum_{t=0}^{k-1}\alpha_t}$. Hence, 
\begin{align}
    \min_{t=0,\dots,k-1} \ec{\sqn{\nabla \phi(\theta_t)}} = \cO\br{\frac{\kappa\log(k)}{\sqrt{k+1}}}.
\end{align}
\end{proof}

\subsubsection{Convergence in expectation for a given precision.}
\begin{proof}
In this proof, we procede as in the proof of Corollary \ref{cor:rsgda_fixed_stepsizes}. For simplicity, we consider that $\sigma^2 \geq 1$. Otherwise, we can simply replace $\sigma$ by $1$, and the proof will still hold with $1$ instead of $\sigma$.

Using $\tilde{\sigma}^2 = 0$ and $\eta_k = \frac{1}{2L}$ in \eqref{eq:inequality_after_telescoping_fixed_step_sizes}, we have
\begin{align}
    \min_{t=0,\dots,k-1} \ecn{\nabla \phi(\theta_t)} \leq \frac{2\cD_0}{\alpha p k} + \frac{4\kappa L^2 r_0}{(1-p) k} + 2\alpha\kappa L \sigma^2 + \frac{8\alpha^2\kappa^4L^2p^2\sigma^2}{(1-p)^2} + \frac{4\alpha^2\kappa^3 L^2p\sigma^2}{(1-p)}.
\end{align}

Let $\epsilon > 0$. The proof follows simply from forcing each term of the LHS of the previous inequality to be smaller than $\frac{\epsilon^2}{5}$.
This results in the following choice of $\alpha$,
\begin{align}
    \alpha = \min\left\{\frac{\epsilon^2}{10\kappa L \sigma^2},\, \frac{(1-p)\epsilon}{2\sqrt{10}\kappa^2 L p \sigma},\, \sqrt{\frac{1-p}{p}}\frac{\epsilon^2}{2\sqrt{5}\kappa\sqrt{\kappa}L\sigma},\, \frac{1-p}{8\kappa^2L\sqrt{p\br{2p + \frac{1-p}{\kappa}}}}\right\},
\end{align}
and the following lower bound on $k$:
\begin{align}
    k \geq \max\left\{\frac{100\cD_0 \kappa L \sigma^2}{p\epsilon^2}, \, \frac{20\sqrt{10}\cD_0\kappa^2 L \sigma}{(1-p)\epsilon^3}, \, \frac{20\sqrt{5}\cD_0\kappa \sqrt{\kappa}L\sigma}{\sqrt{p(1-p)}\epsilon^4}, \, \frac{80\cD_0\kappa^2 L}{(1-p)\epsilon^2\sqrt{2+\frac{1-p}{p\kappa}}}, \, \frac{20\kappa L^2r_0}{(1-p)\epsilon^2}. \right\}
\end{align}
By choosing $p \in \left[\frac{1}{\kappa}, \frac{1}{2}\right]$, we have that the RHS is at most of the order $\Theta\br{\kappa^2\epsilon^{-4}}$.
\end{proof}

\section{Learning with a semi-discrete Sinkhorn loss is a nonconvex-strongly concave problem}\label{sec:app_semi_discrete_sinkhorn}
The goal of this Section is to present the details and proof of Lemma \ref{lem:sinkhorn_strongly_concave}. Let us first recall the problem setting presented in \cref{sec:applications_sinkhorn} in more detail.

\subsection{Problem setting}
Let $\cX \subseteq \R^p$. For all $n \in \N^*$, define $[n] \eqdef \left\{1,\dots,n\right\}$. Assume that we are given a fixed dataset $(y_1,\dots,y_n) \subset \cX$ and that $\nu = \sum_{j=1}^n \bnu_j \delta_{y_j}$, where $\sum_{j=1}^n \bnu_j = 1$ (we generalize the setting of \cref{sec:applications_sinkhorn} to non-uniform probabilities). Then, from \cite{genevay2016}, we have that for any $\mu \in \cP(\cX)$,
\begin{align}
\cW(\mu, \nu) = \max_{v \in \R^n} \eczmu{h(z, v)},
\end{align}
where
\begin{align}\label{eq:h(x,v)}
h(x, v) = \sum_{j=1}^n v_j \bnu_j - \epsilon \log\br{\sum_{i=1}^n \exp\br{\frac{v_i - c(x, y_i)}{\epsilon}}\bnu_i} + \epsilon.
\end{align}

For a mapping $g(\theta \,,\cdot) : \cZ \rightarrow \cX$, we have
\begin{align}
\cW(g(\cdot \,, \theta)_{\#}\mu, \nu) = \max_{v \in \R^n} \eczmu{h\br{g(\theta, Z), v}}.
\end{align}
Thus, the problem of learning with a Sinkhorn loss can be formulated as
\begin{align}\label{eq:semi_dual_problem}
\min_{\theta \in \R^d} \max_{v \in \R^n}\, F(\theta, v) \eqdef \eczmu{h\br{g_z(\theta), v}}.
\end{align}

\noindent Assuming that a minimum exists, our goal is to find
\begin{align}\label{eq:sinkhorn_true_objective}
\theta^* \in \argmin_{\theta \in \R^d} \phi(\theta), \quad \mbox{where} \quad \phi(\theta) = \max_{v \in \R^n} F(\theta, v).
\end{align}

\subsection{Assumptions and consequences}\label{sec:asm_and_csq}
We now detail the assumptions of Lemma \ref{lem:sinkhorn_strongly_concave}.

\begin{assumption}\label{asm:cost_lip_smooth}
We have
\begin{enumerate}
    \item for all $z \in \cZ, \theta \mapsto c(g_z(\theta), y)$ and $y \mapsto c(g_z(\theta), y)$ are $L_c$-lipschitz for $\norm{\cdot}_2$ and $\norm{\cdot}_{\infty}$ respectively \textit{a.s.}
    \item for all $z \in \cZ$ and $y \in \cY$, $\theta \mapsto c(g_z(\theta), y)$ is twice differentiable and $\cL_c$-smooth \emph{a.s.}
\end{enumerate}
\end{assumption}

Assumption \ref{asm:cost_lip_smooth} has the following consequences on the function $F$ defined in \eqref{eq:semi_dual_problem}.
\begin{lemma}\label{lem:sinkhorn_F_lipschitz_smooth}
$v \mapsto F(\theta, v)$ is $\frac{1}{\epsilon}$-smooth for all $\theta \in \R^d$. Moreover, if assumption \ref{asm:cost_lip_smooth} holds, then
\begin{enumerate}
    \item $\theta \mapsto F(\theta, v)$ is $L_c$-lipschitz and $L_F$-smooth, where $L_F \eqdef \frac{\cL_c + 2L_c^2}{\epsilon}$.
    \item $v \mapsto \nabla_\theta F(\theta, v)$ and $\theta \mapsto \nabla_v F(\theta, v)$ are $\frac{2L_c}{\epsilon}$-lipschitz.
\end{enumerate}
\end{lemma}

Before proving Lemma \ref{lem:sinkhorn_F_lipschitz_smooth}, we present some notations and preliminary calculations that will be used throughout the proofs. 
\paragraph{Notations.} We define for all $\theta \in \R^d$, $v \in \R^n$, $z \in \cZ$, $x, y \in \cX^2$
\begin{align}
    c_j(x) &\eqdef c(x, y_j)\\
    w_j(x) &\eqdef \frac{\exp\br{\frac{v_j - c_j(x)}{\epsilon}}\bnu_j}{\sum_{k=1}^n \exp\br{\frac{v_k - c_k(x)}{\epsilon}}\bnu_k}\\
    c(g_z(\theta),y) &= \left[c_1(g_z(\theta)), \dots, c_n(g_z(\theta))\right]^\top 
\end{align}

\paragraph{Closed form gradients.} Let  $\theta \in \R^d$, $v \in \R^n$, $z \in \cZ$ and $y \in \cX$. We have
\begin{align}
\nabla_v h(g_z(\theta), v) = \bnu - \frac{\exp\br{\frac{v - c(g_z(\theta),y)}{\epsilon}} \odot \bnu}{\exp\br{\frac{v - c(g_z(\theta),y)}{\epsilon}}^\top \bnu} \in \R^n, \label{eq:gradient_v_h}
\end{align}
and
\begin{align}
\nabla_\theta h(g_z(\theta), v) =  \sum_{j=1}^n\frac{\exp\br{\frac{v_j - c_j(g_z(\theta))}{\epsilon}}\bnu_j}{\sum_{k=1}^n \exp\br{\frac{v_k - c_k(g_z(\theta))}{\epsilon}}\bnu_k}\nabla_\theta c_j(g_z(\theta)) \in \R^d.
\end{align}

\begin{proof}[Proof of Lemma \ref{lem:sinkhorn_F_lipschitz_smooth}]\label{sec:app_proof_lemma_smoothness}
Let Assumption \ref{asm:cost_lip_smooth} hold. Let $y \in \cX$.
\begin{itemize}[leftmargin=*]
\item \textbf{$\mathbf{v \mapsto F(\theta, v)}$ is $\mathbf{\frac{1}{\epsilon}}$-smooth.}

Let  $\theta \in \R^d$, $v \in \R^n$, $z \in \cZ$. Define $a \eqdef \exp\br{\frac{v - c(g_z(\theta), y)}{\epsilon}} \odot \bnu$. Differentiating $v \mapsto h(g_z(\theta), v)$ twice, we have
\begin{align}
    \nabla_v^2 h(g_z(\theta), v) = \frac{1}{\epsilon}\br{\frac{aa^\top}{(a^\top \ones_n)^2} - \frac{\diag(a)}{a^\top \ones}}
\end{align}
Using the Cauchy-Schwartz inequality, we can show that $\nabla_v^2 h(x, v) \preceq 0$. Moreover, we have for all $b \in \R^n$,
\begin{align}
    \frac{b^\top \diag(a) b}{a^\top \ones} = \sum_{i=1}^n\frac{a_i }{\sum_{j=1}^n a_j}b_i^2 \leq \sqn{b}, 
\end{align}
that is, $\frac{\diag(a)}{a^\top \ones} \succeq I$. This, together with the fact that $\frac{aa^\top}{(a^\top \ones_n)^2} \succeq 0$, implies that $-\frac{1}{\epsilon} I \preceq \nabla_v^2 h(x, v) \preceq 0$. This in turn implies that $v \mapsto F(\theta, v)$ is $\frac{1}{\epsilon}$-smooth.

\item \textbf{$\mathbf{\theta \mapsto F(\theta, v)}$ is $\mathbf{\frac{\cL_c + 2L_c^2}{\epsilon}}$-smooth.}

Let  $\theta \in \R^d$, $v \in \R^n$, $z \in \cZ$. We have
\begin{align}
    \nabla_\theta^2 h(g_z(\theta), v) & = \frac{1}{\epsilon} \sum_{j=1}^n w_j(g_z(\theta)) \br{\nabla^2 c_j(g_z(\theta)) - \nabla c_j(g_z(\theta)) \nabla c_j(g_z(\theta))^\top}\sum_{k=1}^n w_k(g_z(\theta))\nabla_\theta c_k(g_z(\theta)) \\
    &+ \br{\sum_{j=1}^n w_j(g_z(\theta))\nabla_\theta c_j(g_z(\theta))}\br{\sum_{k=1}^n w_k(g_z(\theta))\nabla_\theta c_k(g_z(\theta))^\top}\\
    &= \frac{1}{\epsilon} \sum_{k=1}^n w_k(g_z(\theta)) \sum_{j=1}^n w_j(g_z(\theta)) \br{\nabla_\theta^2 c_j(g_z(\theta)) + \br{\nabla_\theta c_k(g_z(\theta)) - \nabla_\theta c_j(g_z(\theta))}\nabla_{\theta} c_j(g_z(\theta))^\top} 
\end{align}
Let $b \in \R^d$. Then
\begin{align}
    \norm{\nabla_\theta^2 h(g_z(\theta), v)b} &\leq \frac{1}{\epsilon} \sum_{k=1}^n w_k(g_z(\theta)) \sum_{j=1}^n w_j(g_z(\theta)) \br{\norm{\nabla_\theta^2 c_j(g_z(\theta))b} +  \norm{\nabla_\theta c_k(g_z(\theta)) - \nabla_\theta c_j(g_z(\theta))}\abs{\nabla_{\theta} c_j(g_z(\theta))^\top b}}
\end{align}
Using Cauchy-Schwarz and the fact that $\theta \mapsto c_j(g_z(\theta))$ is $L_c$-lipschitz, we have $$\abs{\nabla_{\theta} c_j(g_z(\theta))^\top b} \leq \norm{\nabla_{\theta} c_j(g_z(\theta))}\norm{b} \leq L_c \norm{b} \quad \mbox{and} \quad \norm{\nabla_\theta c_k(g_z(\theta)) - \nabla_\theta c_j(g_z(\theta))} \leq 2L_c.$$ Moreover, since $\theta \mapsto c_j(g_z(\theta))$ is $\cL_c$-smooth, we have $\norm{\nabla_\theta^2 c_j(g_z(\theta))b} \leq \cL_c \norm{b}$.
Hence,
\begin{align}
    \norm{\nabla_\theta^2 h(g_z(\theta), v)b} &\leq \frac{2L_c^2 + \cL_c}{\epsilon}\norm{b} \sum_{k=1}^n w_k(g_z(\theta)) \sum_{j=1}^n w_j(g_z(\theta))\\
    &= \frac{2L_c^2 + \cL_c}{\epsilon}\norm{b}.
\end{align}
Taking the expectation and using Jensen's inequality gives 
\begin{align}
    \norm{\nabla_\theta^2 F(\theta, v)b} \leq \eczmu{\norm{\nabla_\theta^2 h(g_z(\theta), v)b}} \leq \frac{2L_c^2 + \cL_c}{\epsilon}\norm{b}.
\end{align}
This shows that $\theta \mapsto F(\theta, v)$ is $\frac{2L_c^2 + \cL_c}{\epsilon}$-smooth.

\item \textbf{$\mathbf{v \mapsto \nabla_\theta F(\theta,v)}$ and $\mathbf{\theta \mapsto \nabla_v F(\theta,v)}$ are $\mathbf{\frac{2L_c}{\epsilon}}$-lipschitz.}

Let  $\theta \in \R^d$, $v \in \R^n$, $z \in \cZ$. Let $a \in \R^d$. Differentiating $\theta \mapsto \nabla_v h(g_z(\theta),v)$, we have
\begin{align}
    \sqn{\nabla_{\theta v}^2 h(g_z(\theta),v) a} &= \frac{1}{\epsilon^2} \sum_{i=1}^n \br{w_i(g_z(\theta)) \sum_k w_k(g_z(\theta)) \br{\nabla_\theta c_i(g_z(\theta)) - \nabla_\theta c_k(g_z(\theta))}^\top a  }^2 \\
    &= \frac{1}{\epsilon^2} \sum_{i=1}^n w_i(g_z(\theta))^2\br{\sum_k w_k(g_z(\theta)) \br{\nabla_\theta c_i(g_z(\theta)) - \nabla_\theta c_k(g_z(\theta))}^\top a  }^2 \\
    &\leq \frac{1}{\epsilon^2} \sum_{i=1}^n w_i(g_z(\theta))^2\sum_k w_k(g_z(\theta)) \br{\br{\nabla_\theta c_i(g_z(\theta)) - \nabla_\theta c_k(g_z(\theta))}^\top a }^2 \\
    &\leq \frac{1}{\epsilon^2} \sum_{i=1}^n w_i(g_z(\theta))^2\sum_k w_k(g_z(\theta)) \sqn{\nabla_\theta c_i(g_z(\theta)) - \nabla_\theta c_k(g_z(\theta))} \sqn{a}\\
    &\leq \frac{4L_c^2}{\epsilon^2}  \sqn{a} \sum_{i=1}^n w_i(\theta)^2\\
    &\leq \frac{4L_c^2}{\epsilon^2} \sqn{a},
\end{align}
where we used Jensen's inequality in the first inequality, Cauchy-Schwarz in the second, the fact that $\theta \mapsto c_i(\theta)$ is $L_c$-lipschitz \emph{a.s.} in the third, and $w_i(\theta)^2 \leq w_i(\theta)$ in the fourth. Note that the norm on the left hand side is in $\R^n$, while that on the right hand side is in $\R^d$. 
Since the squared norm is convex, using Jensen's inequality,
\begin{align}
    \sqn{\nabla_{\theta v}^2 F(\theta,v) a} \leq \eczmu{\sqn{\nabla_{\theta v}^2 h(g_z(\theta),v) a}} \leq \frac{4L_c^2}{\epsilon^2} \sqn{a}.
\end{align}
We conclude that $\theta \mapsto \nabla_vF(\theta, v)$ is $\frac{2L_c}{\epsilon}$-lipschitz for all $v \in \R^n$. Similarly, we have that $v \mapsto \nabla_\theta F(\theta, v)$ is $\frac{2L_c}{\epsilon}$-lipschitz for all $\theta \in \R^d$.
\end{itemize}
\end{proof}

\subsection{Restricting the maximization problem}
In this section, we give the explicit form of the set $\cV$ in Lemma \ref{lem:sinkhorn_strongly_concave}. Indeed, we show that the problem $\underset{v \in \R^n}{\max}\, F(\theta, v)$ can be restricted to a smaller bounded set on which $v \mapsto F(\theta, v)$ is strongly-concave for all $\theta \in \R^d$.
\begin{lemma}
Let Assumption \ref{asm:cost_lip_smooth} hold. Define
\begin{align}\label{eq:def_V}
    \cV \eqdef \left\{v \in \R^n: \; \norm{v}_2 \leq \frac{L_c}{n} \sum_{i, k = 1}^n \norm{y_k - y_i}_\infty \, \text{and}\; \sum_{i=1}^n v_i = 0 \right\}.
\end{align}
Then, for all $\theta \in \R^d$,
\begin{align}\label{eq:sinkhorn_max_problem}
    \max_{v \in \R^n} F(\theta, v) = \max_{v \in \cV} F(\theta, v),
\end{align}
and $F$ admits a unique maximizer on $\cV$.
\end{lemma}

\begin{proof}
Let $v^* \in \argmax_{v \in \R^n} F(\theta, v)$. Since $\theta \mapsto  c(g_z(\theta), y)$ is $L_c$-lipschitz, we have that the (Sinkhorn) Kantorvitch potentials are $L_c$-Lipschitz on $\cY$ as well \citep[Proposition 1]{genevay2019sample}. This means, in the discrete setting, that for any $v^* \in \argmax_{v \in \R^n} F(\theta, v)$ and $(k, i) \in [n]^2$,
\begin{align}
    \abs{v^*_k - v^*_i} \leq L_c \supnorm{y_k - y_i}.
\end{align}
Hence,
\begin{align}
    v^*_i - L_c \supnorm{y_k - y_i} \leq v^*_k \leq v^*_i + L_c \supnorm{y_k - y_i}.
\end{align}
Further, since the Kantorovitch potentials are uniquely defined upto a constant, we can restrict our attention to the \textit{unique} vector $v^* \in \R^n$ which verifies $\sum_{i=1}^n v^*_i = 0$. Thus, summing the previous inequality from $i=1$ to $n$ and rearranging, we have
\begin{align}
    \abs{v^*_k} \leq \frac{L_c}{n} \sum_{i=1}^n \supnorm{y_k - y_i}.
\end{align}
Finally, summing between $k=1$ and $n$ and using the fact that $\norm{v}_2 \leq \norm{v}_1$, we have that
\begin{align}
    \norm{v^*}_2 \leq \frac{L_c}{n} \sum_{i,k=1}^n \supnorm{y_k - y_i}.
\end{align}
That is, the unique solution $v^*$ to the maximization problem $\underset{v \in \R^n}{\max} F(\theta, v)$ which verifies $\sum_{i=1}^n v^*_i = 0$ also verifies the previous inequality. Thus, we can restrict the maximization to the set $\cV$ defined in \eqref{eq:def_V}
\end{proof}

\begin{remark}
Suppose we want to solve Problem \eqref{eq:sinkhorn_max_problem} using Projected Stochastic Gradient Ascent. For simplicity, suppose here that $g(z, \theta) = z$ for all $z, \theta \in \cZ \times \R^d$ \textit{almost surely}. Suppose that we initialize at $v_0 = 0$ and update for all $k \in \N$, $v_{k+1} = \Pi_{\cV}\br{v_k + \eta \nabla_v h(z_k, v_k)}$, where $z_k \sim \mu$. From \eqref{eq:gradient_v_h}, we have that for all $z \in \cZ$ and $v \in \R^n, \; \sum_{j=1}^n \partial_{v^{j}} h(z, v) = 0$, so that, since we initialize at $v_0 = 0$, we will have $\sum_{j=1}^n v_k^{j} = 0$ for all $k \in \N$. Hence, we can replace the set $\cV$ by $\tilde{\cV}  \eqdef \left\{v \in \R^n: \; \norm{v}_2 \leq \frac{L_c}{n} \sum_{i, k = 1}^n \norm{y_k - y_i}_\infty\right\}.$ Let $\beta = \frac{L_c}{n} \sum_{i,k=1}^n \supnorm{y_k - y_i}.$ The projection operator is then simply
\begin{align}
    \Pi_{\tilde{\cV}}(v) = \frac{\beta v }{\max\left\{\norm{v}, \beta\right\}}
\end{align}
for all $v \in \R^n$.
\end{remark}

\subsection{The semi-dual objective is strongly concave on the subset $\cV$}
We now show that $v \mapsto F(\theta, v)$ is strongly concave on $\cV$ for all $\theta \in \R^d$.

\begin{proposition}
Let Assumption \ref{asm:cost_lip_smooth} hold. Define
\begin{align}
    \cV \eqdef \left\{v \in \R^n: \; \norm{v}_2 \leq \frac{L_c}{n} \sum_{i, k = 1}^n \norm{y_k - y_i}_\infty \, \text{and}\; \sum_{i=1}^n v_i = 0 \right\}.
\end{align}
Let $\Delta_y^n = \max_{i, k \in [n]} \norm{y_k - y_i}_{\infty}$. Then, the function $v \mapsto F(\theta, v)$, where $F$ is defined in \eqref{eq:semi_dual_problem}, is $\xi$-strongly concave on $\cV$, with
$$\xi \eqdef \frac{\exp\br{\frac{-2(n+2)L_c\Delta_y^n}{\epsilon}}\min_{k \in [n]}\bnu_k}{2n\epsilon}.$$
\end{proposition}

\begin{proof}
Let $v, d \in \cV$ and define $a = \exp(\frac{v - c(g_z(\theta), y)}{\epsilon}) \odot \bnu$. By differentiating $v \mapsto h(g_z(\theta), v)$ twice, we have, using standard computations for the log-sum-exp function:
\begin{align}
    \nabla_v^2 h(x, v) = \frac{1}{\epsilon}\br{ \frac{aa^\top}{(\ones_n\top a)^2} - \frac{\diag(a)}{\ones_n\top a}}.
\end{align}
Hence,
\begin{align}
d^\top \nabla_v^2 h(x, v)d &= \frac{1}{\epsilon}\br{\frac{(\sum_{i = 1}^n a_i d_i)^2 }{(\ones_n\top a)^2} - \sum_{i=1}^n a_i d_i^2}\\
&= \frac{1}{\epsilon}\br{\frac{(\sum_{i = 1}^n a_i d_i)^2 }{(\ones_n\top a)^2} - \frac{\sum_{i=1}^n a_i d_i^2}{\ones_n \top a}}\\
\end{align}
Define $w_i = \frac{a_i}{\sum_{k=1}^{n}a_k}$. Then, we can rewrite the previous equality as
\begin{align}
d^\top \nabla_v^2 h(x, v)d  &= \frac{1}{\epsilon}\br{(\sum_{i = 1}^n w_i d_i)^2  - \sum_{i=1}^n w_i d_i^2}
\end{align}
Since $\sum_{j=1}^n w_j = 1$, it is easy to show that
\begin{align}
    \br{\sum_{i=1}^n w_i d_i^2 -  (\sum_{i = 1}^n w_i d_i)^2} = \frac{1}{2} \sum_{ij}w_i w_j \br{d_i - d_j}^2.
\end{align}
Hence,
\begin{align}
d^\top \nabla_v^2 h(x, v)d &= -\frac{1}{2\epsilon}\br{\sum_{ij} w_i w_j \br{d_i - d_j}^2}.
\end{align}
But we have for all $i \in [n],$ $ w_i = \frac{\exp\br{\frac{v_i}{\epsilon}}\bnu_i}{\sum_{k=1}^n \exp\br{\frac{v_k + c(x, y_i) - c(x, y_k)}{\epsilon}}\bnu_k}$. And since $y \mapsto c(x, y)$ is $L_c$-lipschitz for all $x \in \cX$, we have 
\begin{align}\label{eq:lip-bound-c}
\abs{c(x, y_i) - c(x, y_k)} \leq L_c \supnorm{y_i - y_k} \leq L_c \Delta_y^n, \quad \mbox{where} \quad \Delta_y^n = \max_{i, k \in [n]} \supnorm{y_i - y_k}.    
\end{align}
Since $v \in \cV$, we also have that $$\norm{v}_2 \leq \frac{L_c}{n}\sum_{i, k}\norm{y_k - y_i}_{\infty} \leq n L_c \Delta_y^n.$$
Using these two inequalities to lower bound $w_i$, we have that
\begin{align}
    w_i \geq \frac{\exp\br{\frac{-(n+2)L_c\Delta_y^n}{\epsilon}}\min_{k \in [n]}\bnu_k}{n}.
\end{align}
Hence,
\begin{align}
d^\top \nabla_v^2 h(x, v)d \leq -\frac{\exp\br{\frac{-2(n+2)L_c\Delta_y^n}{\epsilon}}\min_{k \in [n]}\bnu_k}{2n^2\epsilon}\sum_{ij} \br{d_i - d_j}^2.
\end{align}
But since $d \in \cV$, we have that $\sum_{i=1}^n d_i = 0$. Hence; $\sum_{ij}(d_i - d_j)^2 = 2n\sum_{i=1}^n d_i^2 = 2n \sqn{d}$.
Hence,
\begin{align}
d^\top \nabla_v^2 h(g_z(\theta), v)d \leq -\frac{\exp\br{\frac{-2(n+2)L_c\Delta_y^n}{\epsilon}}\min_{k \in [n]}\bnu_k}{2n\epsilon}\sqn{d}.
\end{align}
This inequality holds for all $z \sim \cD$ almost surely. Taking the expectation shows that $v \mapsto F(\theta, v)$ is strongly concave on $\cV$ for all $\theta \in \R^d$.
\end{proof}

\section{Details about the experimental setting of Section \ref{sec:applications_sinkhorn}}\label{sec:app_experiments_details}

For ease of exposition, we redefine here the semi-dual function used in \cref{sec:applications_sinkhorn}. Let $x \in \cX$, and consider a point cloud $\br{y_j}_{j=1}^n \subseteq \cY$. Then we define for all $v \in \R^n$,
\begin{align}\label{eq:ot_semidual_appendix}
    h\br{x, \br{y_j}_{j=1}^n; v} \eqdef \frac{1}{n} \sum_{j=1}^n v_j - \epsilon \log\br{\frac{1}{n} \sum_{j=1}^n \exp\br{\frac{v_j - c(x, y_j)}{\epsilon}}} - \epsilon.
\end{align}
The goal is to solve the following problem
\begin{align}
    \min_{\theta \in \R^d} \cW_{\epsilon}\br{\mu, \sum_{j=1}^n \delta_{f_\theta(y_j)}} \equiv \max_{v \in \R^n} \; \EE{x \sim \mu}{h\br{x, \br{f_\theta(y_j)}_{j=1}^n; v}}.
\end{align}
Thus, if $\mu = \frac{1}{m}\sum_{i=1}^m \delta_{x_i}$ for $\br{x_i}_{i=1}^m \subseteq \cX$, we want to solve
\begin{align}
    \min_{\theta \in \R^d} \max_{v \in \R^n} \; \frac{1}{m} \sum_{i=1}^m h\br{x_i, \br{f_\theta(y_j)}_{j=1}^n; v}.
\end{align}

\subsection{Dataset, architecture and OT hyperparameters}

\textbf{Dataset.} The dataset we used was first considered in \cite{stark2020scim}. It consists of a single-cell profile of a metastatic melanoma sample from the Tumor Profiler Consortium\footnote{https://tpreports.nexus.ethz.ch/download/scim/data/tupro/}. It contains single-cell data from a cohort of patients which is analyzed using two technologies: Cytometry by Time of Flight (CyTOF, \cite{bandura2009mass}) and scRNA-sequencing \citep{tang2009mrna}. The number of cells analyzed using CyTOF was $m=135334$ and the dimension of the resulting points is $d=41$, and the number of cells analyzed using scRNA-sequencing was $n=4683$, with a dimension $d^\prime = 1024$. We denote the CyTOF point cloud by $\br{x_i}_{i=1}^m$ and $\mu = \frac{1}{m}\sum_{i=1}^m \delta_{x_i}$, and the scRNA point cloud by $\br{y_j}_{j=1}^n$ and $\nu = \frac{1}{n}\sum_{j=1}^n \delta_{y_j}$.

\textbf{Architecture.} Instead of mapping both point clouds to a lower dimensional space, we instead map the smaller point cloud $\nu$ to the larger one $\mu$. This strategy has two benefits: \textit{(i)} it reduces the possible error due to learning the parametric map since we only incur the error for one point cloud, \textit{(ii)} it is computationally less expensive, since we need to displace a smaller point cloud instead of both point clouds. The map we used was a 3-layer MLP with a number of hidden units 128-96-64, and we used GELU activation functions \citep{hendrycks2016gaussian}. We used a minibatch size $b=128$.

\textbf{OT hyperparameters.} We used a regularization parameter $\epsilon = 0.1$ and a quadratic cost function ${c(x, y) = \sqn{x - y}}$ for all $x, y \in \cX$.

\subsection{Using Sinkhorn's algorithm}
We now present Sinkhorn's algorithm.

\begin{algorithm}[H]
\caption{\code{Sinkhorn}($\br{x_i}_{i=1}^m, \, \br{y_j}_{j=1}^n, c, \epsilon, m_{sin}, a_0, b_0$)}
\label{alg:sinkhorn}
\begin{algorithmic}
    \State \textbf{Inputs:} point clouds $\br{x_i}_{i=1}^m$, $\br{y_j}_{j=1}^n$, cost function $c$, regulatization $\epsilon$, number of iterations $m$. Optional: initialization $(a_0, b_0) \in \R_{+}^m \times \R_{+}^n$.
    \State Compute $K \in \R^{n \times n}$, where $K_{i,j} = \exp\br{-\frac{c(x_i, y_j)}{\epsilon}}$ for all $(i, j)$
    \State \textbf{Initialisation:} by default $a_0 = \ones_m$, $b_0 = \ones_n$
    \For{$\ell=0,\dots,m_{sin}-1$}
    \State $a_{\ell+1} = \frac{\ones_m}{Kb_\ell}, \; b_{k+1} = \frac{\ones_n}{K^\top a_{\ell+1}}$
    \EndFor
    \State \Return $(a_{m_{sin}}, b_{m_{sin}})$
\end{algorithmic}
\end{algorithm}

In \cref{sec:applications_sinkhorn}, we use Sinkhorn's algorithm (Alg.~\ref{alg:sinkhorn}) as a subroutine instead of the gradient ascent step in RSGDA (Alg.~\ref{alg:RSGDA}). We present the resulting algorithm explicitely in Alg.~\ref{alg:sinkhorn_learning}.

\begin{algorithm}[H]
\caption{Learning using Sinkhorn's algorithm}
\label{alg:sinkhorn_learning}
\begin{algorithmic}
    \State \textbf{Inputs:} $\br{x_i}_{i=1}^m$, $\br{y_j}_{j=1}^n$, cost function $c$, regulatization $\epsilon$, number of iterations $m$, parameteric map $f_{\theta}$. $a_0 = \ones_m, \; b_0 = \ones_n$, $\theta_0 \in \R^d$, minibatch size $b$, step size $\alpha$
    \For{$k=0,\dots, K-1$}
    \State Sample a minibatch $B \subset [n]$ of size $|B| = b$.
    \State $a_{k+1}, b_{k+1} = \code{Sinkhorn}(\br{x_i}_{i \in B}, \br{y_j}_{j=1}^n, c, \epsilon, m_{sin}, a_k, b_k)$ \Comment{Using Alg.~\ref{alg:sinkhorn}}
    \State $v_{k+1} = -\epsilon \log\br{b_{k+1}}$
    \State $\theta_{k+1} = \theta_k - \frac{\alpha}{b}\sum\limits_{i \in B} h\br{x_i, \br{f_\theta(y_j)}_{j=1}^n; v_{k+1}}$ \Comment{$h$ defined in \eqref{eq:ot_semidual_appendix}}
    \EndFor
\end{algorithmic}
\end{algorithm}
Another way to use Sinkhorn's algorithm would be to initialize each algorithm with $a_0 = \ones_m$ and $b_0 = \ones_n$, but the resulting algorithm was not competitive.
\subsubsection{Impact of the choice of the number of iterations of Sinkhorn's algorithm}
Here, we examine how the number of iterations of Sinkhorn's algorithm $m_{sin}$ should be set in Alg.~\ref{alg:sinkhorn_learning}. We find that we should only use one step, and that the more steps we use, the slower the optimization. This is due to the bias introduced by solving a minibatch version of the true transport problem $\cW_\epsilon(\mu, \nu_{\theta_k})$ \eqref{eq:sc_ot_problem} at each iteration. This bias is already known to be an issue in computing Sinkhorn's loss using minibatch approximations~\citep{fatras2020learning}.

\begin{figure}[H]
    \centering
    \includegraphics[scale=0.55]{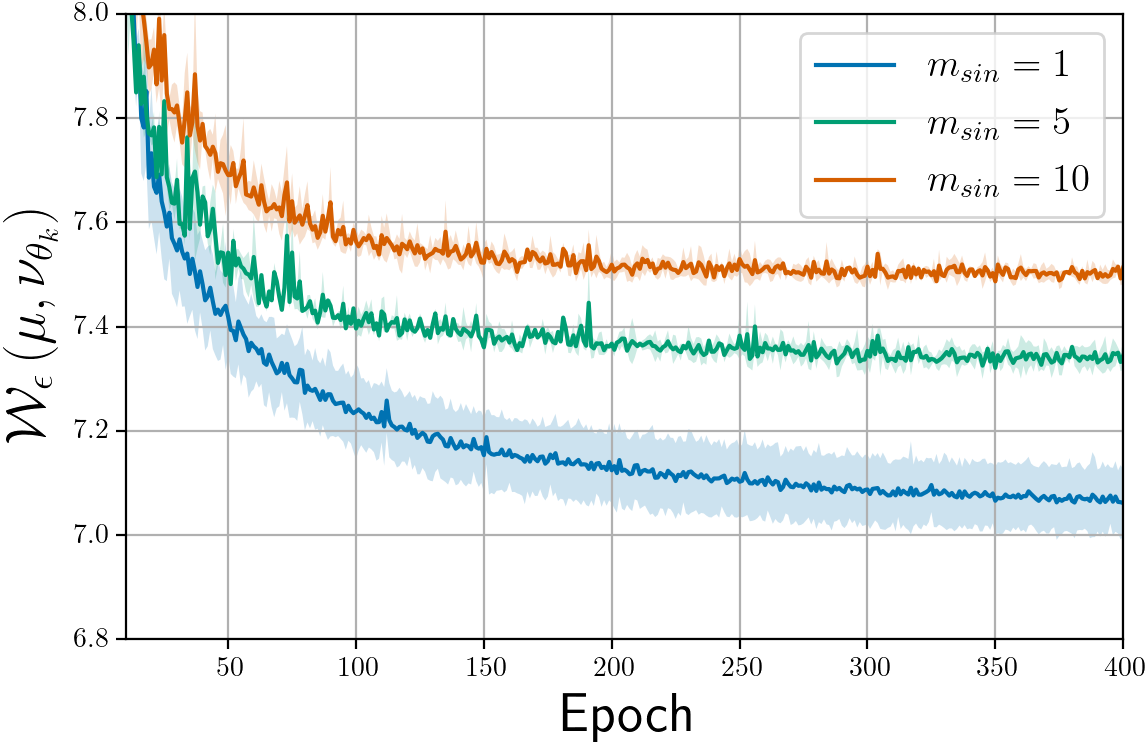}
    \caption{Loss $\cW_{\epsilon}\left(\mu, \nu_{\theta_k}\right)$ \eqref{eq:sc_ot_problem} when using Alg.~\ref{alg:sinkhorn_learning} depending on the number of iterations $m_{\sin}$ of Sinkhorn's algorithm. $\alpha=0.005$, $b = 128$.}
    \label{fig:sc_sinkhorn_comp}
    \vspace{-1.5em}
\end{figure}

\end{appendices}